\documentclass[10pt,twocolumn,letterpaper]{article}

\usepackage[pagenumbers]{cvpr} 

\usepackage{graphicx}
\usepackage{amsmath}
\usepackage{amssymb}
\usepackage{booktabs}
\usepackage{bm}
\usepackage{amsthm}

\usepackage{multirow}
\usepackage[accsupp]{axessibility}

\usepackage[pagebackref,breaklinks,colorlinks]{hyperref}

\usepackage[capitalize]{cleveref}
\crefname{section}{Sec.}{Secs.}
\Crefname{section}{Section}{Sections}
\Crefname{table}{Table}{Tables}
\crefname{table}{Tab.}{Tabs.}

\def\cvprPaperID{6710} 
\def\confName{CVPR}
\def\confYear{2023}

\begin{document}

\title{Label Information Bottleneck for Label Enhancement}

\author{Qinghai Zheng$^{1}$,~ Jihua Zhu$^{2}$\footnotemark[1]\thanks{Corresponding author, E-mail: zhujh@xjtu.edu.cn},~ Haoyu Tang$^{3}$ \\
	$^{1}$College of Computer and Data Science, Fuzhou University, China \\
	$^{2}$School of Software Engineering, Xi'an Jiaotong University, Xi'an, China \\
	$^{3}$School of Software, Shandong University, Jinan, China \\
}
\maketitle

\begin{abstract}
   In this work, we focus on the challenging problem of Label Enhancement (LE), which aims to exactly recover label distributions from logical labels, and present a novel Label Information Bottleneck (LIB) method for LE. For the recovery process of label distributions, the label irrelevant information contained in the dataset may lead to unsatisfactory recovery performance. To address this limitation, we make efforts to excavate the essential label relevant information to improve the recovery performance. Our method formulates the LE problem as the following two joint processes: 1) learning the representation with the essential label relevant information, 2) recovering label distributions based on the learned representation. The label relevant information can be excavated based on the ``bottleneck'' formed by the learned representation. Significantly, both the label relevant information about the label assignments and the label relevant information about the label gaps can be explored in our method. Evaluation experiments conducted on several benchmark label distribution learning datasets verify the effectiveness and competitiveness of LIB. Our source codes are available at \url{https://github.com/qinghai-zheng/LIBLE}
\end{abstract}

\section{Introduction}
\label{sec:intro}

Learning with label ambiguity is important in computer vision and machine learning. Different from the traditional Multi-Label Learning (MLL), which employs multiple logical labels to annotate one instance to address the label ambiguity issue \cite{liu2021emerging}, Label Distribution Learning (LDL) considers the relative importance of different labels and draws much attention in recent years \cite{TKDE2016_LDL_reivew, gao2018age,wang2021label,jia2019facial,TIP2019_LDL_CrowCounting}. By distinguishing the description degrees of all labels, LDL annotates one instance with a label distribution. Therefore, LDL is a more general learning paradigm, MLL can be regarded as a special case of LDL \cite{TKDE2016_LDL_reivew,jia2019labelTKDE,hong2021disentangling}. 

Recently, many LDL methods are proposed and achieve great success in practice \cite{TIP2019_LDL_CrowCounting,geng2015pre,jia2019facial,chen2020label}. Instances with exact label distributions are vital for the training process of LDL methods. Nevertheless, annotating instances with label distributions is time-consuming\cite{xuLETKDE, tang2020label}. We take the label distribution annotation process of SJAFFE dataset for example here. SJAFFE dataset is the facial expression dataset, which contains 213 grayscale images collected from 10 Japanese female models, each facial expression image is rated by 60 persons on 6 basic emotions, including happiness, surprise, sadness, fear, anger, and disgust, with a five-level scale from 1 - 5, the higher value indicates the higher emotion intensity. Consequently, the average score of each emotion is served as the emotion label distribution \cite{xuLETKDE,jia2019facial}. Clearly, the above annotation process is costly and it is unpractical to annotate data with label distributions manually, especially when the number of data is large. Fortunately, most existing datasets in the field of computer vision and machine learning are annotated by single-label or multi-labels \cite{xu2020variational,gao2021label}, therefore, a highly recommended promising solution is Label Enhancement (LE), which attempts to recover the desired label distributions exactly from existing logical labels \cite{tang2020label,xuLETKDE,GLESC}.

Driven by the urgent requirement of obtaining label distributions and the convenience of LE, some LE methods are proposed in recent years~\cite{xuLETKDE,ML_LE,LP_LE,FCM_LE,KM_LE,lv2019weakly,jia2021labelTKDE,xu2020variational,tang2020label,gao2021label}. Given a dataset $\bm{X} = \{ {\bm{x}_1},{\bm{x}_2}, \cdots ,{\bm{x}_n}\}  \in {\mathbb{R}^{q \times n}}$, in which $q$ and $n$ denote the number of dimensions and the number of instances, the potential label set is $\{ {y_1},{y_2}, \cdots ,{y_c}\}$. The available logical labels and the desired distribution labels of ${\bm{X}}$ are separately indicated by $\bm{L} = \{ {\bm{l}_1},{\bm{l}_2}, \cdots ,{\bm{l}_n}\} $ and $\bm{D} = \{ {\bm{d}_1},{\bm{d}_2}, \cdots ,{\bm{d}_n}\}$, where $\bm{l}_i$ and $\bm{d}_i$ are:
\begin{equation}
	\bm{l}_i = ( {l_i^{y_1}},{l_i^{y_2}}, \cdots ,{l_i^{y_c}})^T,
	\bm{d}_i = ( {d_i^{y_1}},{d_i^{y_2}}, \cdots ,{d_i^{y_c}})^T. 
\end{equation}
To be specific, LE aims to recover ${\bm{D}}$ based on the information provided by ${\bm{X}}$ and $\bm{L}$. For most existing LE methods, their objectives can be concisely summarized as follows:
\begin{equation}
	\mathop {\min }\limits_\theta ~\left\| {{f_\theta }(\bm{X}) - \bm{L}} \right\|_F^2 + \gamma reg({f_\theta }(\bm{X})),
	\label{obj_basic_existing_methods}
\end{equation}
in which $\bm{D} = {f_\theta }(\bm{X})$, ${f_\theta }(\cdot)$ indicates the mapping from $\bm{X}$ to $\bm{D}$, $reg(\cdot)$ denotes the regularization function, and $\gamma$ is the trade-off parameter. Most existing LE methods vary in $reg(\cdot)$. For example, GLLE \cite{xuLETKDE} calculates the distance-based similarity matrix of data and employs the smoothness assumption \cite{Smooth_assumption} to construct $reg(\cdot)$; LESC \cite{tang2020label} considers the global sample correlations and introduces the low-rank constraint as the regularization; PNLR \cite{jia2021labelTKDE} leverages $reg(\cdot)$ to maintain positive and negative label relations during the recovery process. Although a remarkable progress can be made by aforementioned methods, they ignore the label irrelevant information contained in $\bm{X}$, which prevents the further improvement of recovery results. For example, in the LE task of recovering facial age label distributions, the label irrelevant information, such as specularities information, cast shadows information, and occlusions information, may result in the incorrect mapping process of ${f_\theta }({\cdot})$ and the unsuitable regularization of $reg(\cdot)$, eventually leads to the unsatisfactory recovery performance. 

\begin{figure}[t]
	\centering
	\includegraphics[width=0.7\linewidth]{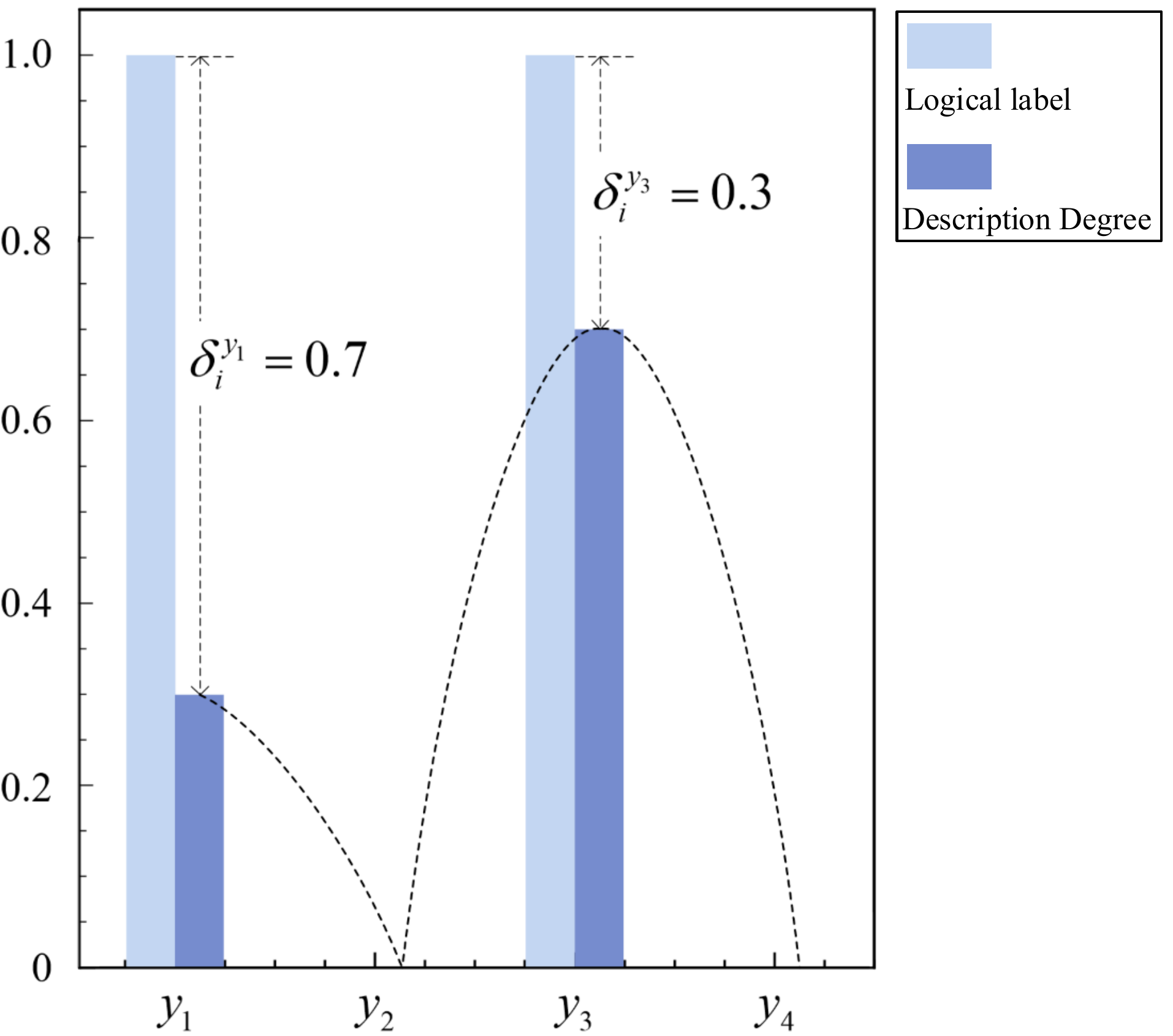}
	\caption{Illustration of label relevant information. Excavating the essential label relevant information directly is challenging and we adopt a indirect way here. We jointly investigate the information about the assignments of labels to the instance and the information about the label gaps between logical labels and label distributions. Given the $i$-th instance $\bm{x}_i$, the label gap of the $y_j$ label is $\delta _i^{{y_j}} = l_i^{{y_j}} - d_i^{{y_j}}$. The information contained in $l_i^{{y_j}}$ and $\delta _i^{{y_j}}$ can be amalgamated to form the essential label relevant information. In other words, we employ $l_i^{{y_j}}$ to explore the label relevant information about the label assignments and $\delta _i^{{y_j}}$ to excavate the label relevant information about the label gaps. To a certain degree, the combination of $\delta _i^{{y_j}}$ and $l_i^{{y_j}}$ is equivalent to $d _i^{{y_j}}$. As depicted here, $l_i^{{y_j}}$ indicates that $y_1$ and $y_3$ are related labels and $\delta _i^{{y_j}}$ provides the importance of $y_1$ and $y_3$.} 
	\label{fig:illustration}
\end{figure}

To overcome the aforementioned limitation, we present a Label Information Bottleneck (LIB) method for LE. Concretely, the core idea of LIB is to learn the latent representation $\bm{H}$, which preserves the maximum label relevant information, from $\bm{X}$, and jointly recovers the label distributions based on the latent representation. For the LE problem, the label relevant information is the information that describes the description degrees of labels. It is tough to explore the label relevant information directly. As shown in Fig.~\ref{fig:illustration}, we decompose the label relevant information into two components, namely the assignments of labels to the instance and the label gaps between label distributions and logical labels. Inspired by Information Bottleneck (IB) \cite{tishby2000information}, LIB utilizes the existing logical labels to explore the information about the assignments of labels to the instance. Unlike simply employing the original IB on the LE task, our method further considers the information about the label gaps between label distributions and logical labels. It is noteworthy that the above two components of the label relevant information are jointly explored in our method, and that is why we term the proposed method Label Information Bottleneck (LIB). The main contributions can be summarized as follows:
\begin{itemize}
	\item[$\bullet$]We decompose the label relevant information into the information about the assignments of labels to instance and the information about the label gaps between logical labels, both of which can be jointly explored during the learning process of our method.
	\item[$\bullet$]We introduce a novel LE method, termed LIB, which excavates the label relevant information to exactly recover the label distributions. Based on the original IB, which explores the label assignments information for LE, LIB further explores the label gaps information.
	\item[$\bullet$]We verify the effectiveness of LIB by performing extensive experiments on several datasets. Experimental results show that the proposed method can achieve the competitive performance, compared to state-of-the-art LE methods.
\end{itemize}


\section{Related Work}
\label{sec:relatedwork}


\subsection{Label Enhancement}

To recover the label distributions from the existing logical labels, many efforts are made recently \cite{xuLETKDE,tang2020label}. In general, most existing LE methods can be roughly divided into two categories, namely, algorithm adaptation and specialized algorithm \cite{TKDE2016_LDL_reivew,tang2020label}.


Algorithm adaptation extends some existing methods to achieve the goal of LE \cite{FCM_LE,KM_LE}. For example, FCM \cite{FCM_LE} recovers the label distributions by utilizing the fuzzy clustering and fuzzy relabeling. To be specific, FCM utilizes the fuzzy C-means clustering to get different clusters and cluster prototypes, then obtains membership degrees of each instance with respect to different cluster prototypes, finally annotates all instances with label distributions by employing the fuzzy composition and softmax normalization. KM \cite{KM_LE} leverages the fuzzy SVM to achieve the membership function. During the recovery process, KM separates instances into two clusters and employs the nonlinear function to get the radius and distances between centers and kernelized instances, and then gets the label distributions with the help of the softmax normalization. 

Specialized algorithm is specially designed to deal with the LE problem. Most existing LE methods belong to the category of specialized algorithm and have the basic objective Eq.~(\ref{obj_basic_existing_methods}). By using different constraints, different methods adopt different $reg(\cdot)$ in Eq.~(\ref{obj_basic_existing_methods}). For example, based on the assumption that instances closed in the feature space are more likely to share the same label, GLLE \cite{xuLETKDE} employs the following local graph information in the feature space to boost the recovery performance:
\begin{equation}
	{{q}_{i,j}}=\left\{ \begin{aligned}
		& \exp \left( -\frac{{{\left\| {{\bm{x}}_{i}}-{{\bm{x}}_{j}} \right\|}^{2}}}{2{{\varepsilon }^{2}}} \right),{\rm{if}}~{\bm{x}}_{j}\in k\left( i \right), \\ 
		& 0,~{\rm{otherwise}}, \\ 
	\end{aligned} \right.
\end{equation}
where $k\left( i \right)$ denotes the $k$-nearest neighbours of ${\bm{x}}_{i}$. $reg(\cdot)$ in GLLE is constructed as follows: 
\begin{equation}
	reg({f_\theta }(\bm{X})) = \sum\limits_{i,j} {{q_{i,j}}\left\| {{f_\theta }({\bm{x}_i}) - {f_\theta }({\bm{x}_j})} \right\|_2^2}. 
\end{equation}
Unlike GLLE, LESC \cite{tang2020label} considers the global graph information and uses the low-rank representation learning \cite{LRR}:
\begin{equation}
	\underset{\bm{G},\bm{E}}{\mathop{\min }}\,{{~\left\| \bm{G} \right\|}_{*}}+{\lambda}_{2} {{\left\| \bm{E} \right\|}_{2,1}}, {~\rm{s.t.}}, \bm{X}=\bm{XG}+\bm{E},
	\label{LESC_objective}
\end{equation}
where  $\bm{G}$ indicates the low-rank representation of instances in the feature space. The regularization function in LESC is written as follows:
\begin{equation}
	reg({f_\theta }(\bm{X})) = \left\| {{f_\theta }({\bm{X}}) - {f_\theta }({\bm{X}})\bm{G}} \right\|_F^2. 
\end{equation}

For these aforementioned LE methods, they all neglect the label irrelevant information contained in $\bm{X}$, the negative effect of which can result in the unsatisfactory recovery results. Taking the recovery of facial emotion label distributions for example, the inaccurate graph information would be obtained in GLLE and LESC with the presence of label irrelevant information, such as the identity information, hindering the further improvement of recovery results.

\subsection{Information Bottleneck}

Information bottleneck (IB) \cite{tishby2000information,tishby2000information,alemi2017deep} is an information theoretic principle, which describes the relevant information in data formally. To be concrete, IB has the following objective:
\begin{equation}
	\mathop {\min }\limits_{\bm{B}}~-I(\bm{B},\bm{C}), {~\rm{s.t.}}, I(\bm{A},\bm{B}) \leqslant I_c,
\end{equation}
where $I(\cdot,\cdot)$ measures the mutual information and $I_c$ is the information constraint. Clearly, IB aims to learn the representation $\bm{B}$, which preserves the relevant information about $\bm{C}$, from $\bm{A}$. Considering the scenario of LE, it is natural to get the following formula:
\begin{equation}
	\mathop {\min }\limits_{\bm{H}}~-I(\bm{H},\bm{L}), {~\rm{s.t.}}, I(\bm{X},\bm{H}) \leqslant I_c.
	\label{sim_IB_LE}
\end{equation}
As discussed in Section \ref{sec:intro}, the information merely about the assignments of labels to the instance can be explored based on Eq.~(\ref{sim_IB_LE}), which neglects the vital information about the label gaps between logical labels and label distributions.

Recently, IB has been successfully utilized in many real-world applications \cite{alemi2017deep,amjad2019learning,wu2020graph,qian2020unsupervised,zhang2021coarse,yu2022improving}. To the best of our knowledge, the method introduced in this paper is the first work that leverages IB to deal with the LE problem. More notably, rather than using IB simply (as shown in Eq.~(\ref{sim_IB_LE})), our method conducts more in-depth exploration to exactly recover label distributions based on IB.  
 
\section{The Proposed Method}
\label{sec:method}


\begin{figure}[t]
	\centering
	\includegraphics[width=0.93\linewidth]{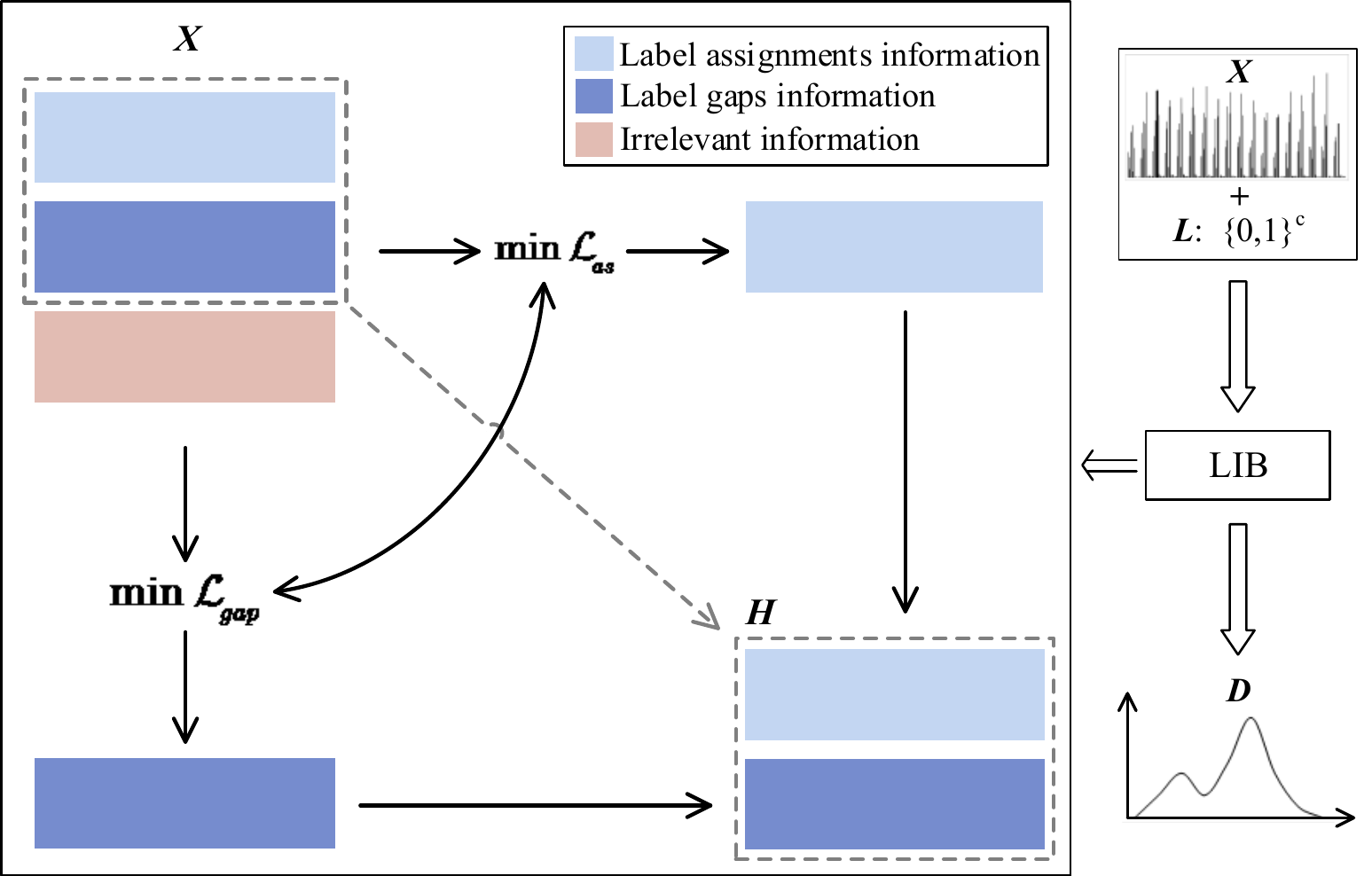}
	\caption{Framework of the proposed LIB. Since the way that directly explores the label relevant information is challenging, LIB decomposes the label relevant information into two components, namely the assignments of labels to the instance and the label gaps between label distributions and logical labels, and adopts an indirect path, which explores the information about label assignments and label gaps simultaneously. By minimizing $\mathcal{L}_{as}$, LIB explores the information about assignments of labels to the instance. Meanwhile, by minimizing $\mathcal{L}_{gap}$, LIB explores the information about the label gaps between logical labels and label distributions. Consequently, the label relevant information can be effectively explored and the label distributions can be exactly recovered.} 
	\label{fig:framework}
\end{figure}

\subsection{The Objective Construction}
Generally, the basic idea can be written as follows:
\begin{eqnarray}
	\mathop {\min }\limits_{\bm{H}}~{{\mathcal{L}}_{as}} + \alpha {{\mathcal{L}}_{gap}},{~\rm{s.t.}},I({\bm{X}},{\bm{H}}) \leqslant {I_c},
	\label{obj:lib_brief}
\end{eqnarray}
where ${{\mathcal{L}}_{as}}$ excavates the information about the \textit{as}signments of labels to the instance, ${{\mathcal{L}}_{gap}}$ investigates the information about the label \textit{gap}s between the logical labels and distribution labels, $\alpha$ is the trade-off parameter, the constraint aims to remove the label irrelevant information. The framework of our method is depicted in Fig.~\ref{fig:framework}. It's worth noting that employing the original IB for LE merely explores the information about the label assignments. While our LIB makes attempts to capture the information about both the label assignments and the description degrees of labels.

\subsubsection{Label assignmens information modeling}

For ${{\mathcal{L}}_{as}}$, inspired by IB, we have the following formula:
\begin{equation}
	{{\mathcal{L}}_{as}} = -I(\bm{H},\bm{L}).
	\label{obj:L_as_brief}
\end{equation}
According to the concept of mutual information, ${{\mathcal{L}}_{as}}$ can be rewritten out in full as follows: 
\begin{equation}
	{{\mathcal{L}}_{as}} =  - \sum\limits_{\bm{h}} {\sum\limits_{\bm{l} } {p(\bm{h},\bm{l})\log \frac{{p(\bm{l}|\bm{h})}}{{p(\bm{l})}}} }. 
\end{equation}
For the convenience of optimization, we introduce the variational approximation $q(\bm{l}|\bm{h})$ to $p(\bm{l}|\bm{h})$. Since both the Kullback Leibler divergence and the entropy are positive:
\begin{equation}
	\begin{gathered}
		{\rm{KL}}({p(\bm{l}|\bm{h})}||{q(\bm{l}|\bm{h})}) = \sum\limits_{\bm{l}} {{p(\bm{l}|\bm{h})}\log \frac{{{p(\bm{l}|\bm{h})}}}{{q(\bm{l}|\bm{h})}}}  \geqslant 0 \hfill \\
		\Rightarrow \sum\limits_{\bm{l}} {{p(\bm{l}|\bm{h})}\log {p(\bm{l}|\bm{h})}}  \geqslant \sum\limits_{\bm{l}} {{p(\bm{l}|\bm{h})}\log {q(\bm{l}|\bm{h})}},  \hfill \\ 
	\end{gathered} 
	\label{kl_pq_1}
\end{equation}
\begin{equation}
	{\mathbb{E}}_{p(\bm{l})}[-\log p(\bm{l})] = - \sum\limits_{\bm{l}} {p(\bm{l}) \log p(\bm{l})} \geqslant 0,
\end{equation}
based on Markov chain that ${\bm{L}}\leftarrow{\bm{X}}\rightarrow{\bm{H}}$, we can get:
\begin{equation}
	{L_{as}} \leqslant  - \sum\limits_{\bm{x}} {\sum\limits_{\bm{l}} {\sum\limits_{\bm{h}} {p({\bm{x}},{\bm{l}})p({\bm{h}}|{\bm{x}})\log q({\bm{l}}|{\bm{h}})} } }.
	\label{obj:label_assignment_bound}
\end{equation}

\subsubsection{Label gaps information modeling}

To investigate the label-relevant information about the description degrees of labels, we introduce the label gaps between logical labels and label distributions $\bm{\Delta}$, and consider the conditional self-information, i.e., $I(\bm{\Delta}|\bm{H})$. Therefore, we construct ${{\mathcal{L}}_{gap}}$\footnote{It can be also interpreted and derived from the view of the probability distribution: $\mathop {\max }\limits_{\bm{\Delta}}~\log p(\bm{H},\bm{\Delta}) \Rightarrow \mathop {\max }\limits_{\bm{\Delta}}~\log p(\bm{\Delta} |\bm{H}) + \log p(\bm{H}) \Rightarrow \mathop {\max }\limits_{\bm{\Delta}}~\log p(\bm{\Delta} |\bm{H})$. We appreciate reviewers for their helpful comments.} as follows:
\begin{equation}
	\begin{gathered}
		{{\mathcal{L}}_{gap}} = I(\bm{\Delta}|\bm{H}) = -\log p(\bm{\Delta} |\bm{H})\hfill \\
		~~~~~~~~ = - \sum\limits_{\bm{\delta}}  {\sum\limits_{\bm{h}} {\log p(\bm{\delta} |\bm{h})} }  \hfill \\
		~~~~~~~~ = - \sum\limits_{\bm{l}} {\sum\limits_{\bm{h}} {\log p(\bm{l} - \bm{\hat{d}}|\bm{h})} }. \hfill \\ 
	\end{gathered} 
	\label{obj:label_gaps_bound}
\end{equation}
where ${\bm{\delta}} = \bm{l} - \bm{\hat{d}}$, $\bm{\hat{d}}$ is the label distribution recoveried in our method.


\subsubsection{Label irrelevant information modeling}
Regarding the label irrelevant information, LIB employs the constraint in Eq.~(\ref{obj:lib_brief}) to discard it during the learning process. $I(\bm{X},\bm{H})$ can be formulated as follows:
\begin{equation}
	I(\bm{X},\bm{H}) =   \sum\limits_{\bm{x}} {\sum\limits_{\bm{h} } {p(\bm{x},\bm{h})\log \frac{{p(\bm{h}|\bm{x})}}{{p(\bm{h})}}} }. 
\end{equation}
Since it is difficult to calculate $p(\bm{h})$ directly, we also introduce the variational approximation $q(\bm{h})$ to $p(\bm{h})$. Similar to Eq.~(\ref{kl_pq_1}), based on ${\rm{KL}}({p(\bm{h})}||{q(\bm{h})}) \geqslant 0$, we have:
\begin{equation}
	\sum\limits_{\bm{h}} {{p(\bm{h})}\log {p(\bm{h})}}  \geqslant \sum\limits_{\bm{h}} {{p(\bm{h})}\log {q(\bm{h})}}.
\end{equation}
Subsequently, the following formula can be written:
\begin{equation}
	\begin{gathered}
	I(\bm{X},\bm{H}) \leqslant \sum\limits_{\bm{x}} {\sum\limits_{\bm{h} } {p(\bm{x},\bm{h})\log \frac{{p(\bm{h}|\bm{x})}}{{q(\bm{h})}}} }  \hfill \\ 
	~~~~~~~~~~~~~~~= \sum\limits_{\bm{x}}{ {\sum\limits_{\bm{l}} {p({\bm{x}},{\bm{l}})}} {\rm{KL}}(p({\bm{h}}|{\bm{x}})||q({\bm{h}}))}. 
	\label{obj_irr_last}
	\end{gathered} 
\end{equation}

\subsubsection{Objective of LIB}

By employing the Lagrange multiplier method and combining Eq.~(\ref{obj:lib_brief}), (\ref{obj:label_assignment_bound}), (\ref{obj:label_gaps_bound}), and (\ref{obj_irr_last}), we have:
\begin{equation}
	\begin{gathered}
		{\mathcal{L}} = {{\mathcal{L}}_{as}} + \alpha {{\mathcal{L}}_{gap}} + \beta I(\bm{X},\bm{H}) \hfill \\
		~~~\leqslant - \sum\limits_{\bm{x}} {\sum\limits_{\bm{l}} {\sum\limits_{\bm{h}} {p({\bm{x}},{\bm{l}})p({\bm{h}}|{\bm{x}},{\bm{l}})\log q({\bm{l}}|{\bm{h}})} } } \hfill \\
		~~~~~~~- \alpha \sum\limits_{\bm{l}} {\sum\limits_{\bm{h}} {\log p(\bm{l} - \bm{\hat{d}}|\bm{h})} } \hfill \\
		~~~~~~~+ \beta \sum\limits_{\bm{x}}{ {\sum\limits_{\bm{l}} {p({\bm{x}},{\bm{l}})}} {\rm{KL}}(p({\bm{h}}|{\bm{x}})||q({\bm{h}}))}. \hfill \\ 
	\end{gathered} 
\end{equation}
where $\beta$ is the Lagrange multiplier. Considering the bound of $\mathcal{L}$ and using the empirical  Monte Carlo approximation of sampling \cite{rezende2014stochastic}, we have the following objective of LIB:
\begin{equation}
	\resizebox{.887\hsize}{!}{$
	\begin{gathered}
		{{\mathcal{L}}_{LIB}} = \frac{1}{n}\sum\limits_{i = 1}^n {[ - \sum\limits_{\bm{h}} {p({\bm{h}}|{\bm{x}_i})\log q({\bm{l}}_i|{\bm{h}})} }  \hfill \\
		+ \beta {\rm{KL}}(p({\bm{h}}|{\bm{x}}_i)||q({\bm{h}}))] - \alpha  \sum\limits_{\bm{l}} {\sum\limits_{\bm{h}} {\log p(\bm{l} - \bm{\hat{d}}|\bm{h})} }. \hfill \\
	\end{gathered}$}
\label{obj_overall}
\end{equation}

\subsection{The Optimization of LIB}

To minimize the objective of ${{\mathcal{L}}_{LIB}}$, we use the reparameterization trick \cite{kingmaautoVAE,rezende2014stochastic}. For $p({\bm{h}}|{\bm{x}})$, we assume that:
\begin{equation}
	p({\bm{h}}|{\bm{x}}) \sim {\mathcal{N}}({{\bm{\mu}} _{{\bm{h}}|{\bm{x}}}},{{\bm{\sigma}} _{{\bm{h}}|{\bm{x}}}^2\bm{I}}),
	\label{opt_start}
\end{equation}	 
where ${{\bm{\mu}} _{{\bm{h}}|{\bm{x}}}}$ and ${{\bm{\sigma}} _{{\bm{h}}|{\bm{x}}}}$ are obtained by using the encoder network $f_{{\theta}_{en}}(\cdot)$, i.e., ${{\bm{\mu}} _{{\bm{h}}|{\bm{x}}}} = f_{{\theta}_{en}}^{{\bm{\mu}}}(\bm{x})$ and ${{\bm{\sigma}} _{{\bm{h}}|{\bm{x}}}} = f_{{\theta}_{en}}^{{\bm{\sigma}}}(\bm{x})$. Subsequently, we have that:
\begin{equation}
	\bm{h} = {{\bm{\mu}} _{{\bm{h}}|{\bm{x}}}} + {{\bm{\sigma}} _{{\bm{h}}|{\bm{x}}}} \odot \bm{\epsilon},
	\label{obj_get_h}
\end{equation}
where $\bm{\epsilon} \sim {\mathcal{N}}({\bm{0}},{\bm{I}})$ and $\odot$ is the element-wise product. For $q({\bm{l}}|{\bm{h}})$, we assume:
\begin{equation}
	q({\bm{l}}|{\bm{h}}) \sim {\mathcal{N}}({{\bm{\mu}} _{{\bm{l}}|{\bm{h}}}},\bm{I}),
\end{equation}
where ${{\bm{\mu}} _{{\bm{l}}|{\bm{h}}}}$ is learned by using the decoder network $f_{{\theta}_{de}}(\cdot)$, namely, ${{\bm{\mu}} _{{\bm{l}}|{\bm{h}}}} = f_{{\theta}_{de}}(\bm{h})$. For $q({\bm{h}})$, we assume that:
\begin{equation}
	q({\bm{h}}) \sim {\mathcal{N}}({\bm{0}},{\bm{I}}).
\end{equation}
For $p(\bm{l} - \bm{\hat{d}}|\bm{h})$, the following assumption is adopted:
\begin{equation}
	p(\bm{l} - \bm{\hat{d}}|{\bm{h}}) \sim {\mathcal{N}}({\bm{0}},{{\bm{\sigma}} _{{\bm{\delta}}|{\bm{h}}}^2\bm{I}}),
\end{equation}
where ${\bm{\sigma}} _{{\bm{\delta}}|{\bm{h}}}$ can be achieved by introducing the gap deviation network $f_{{\theta}_{gd}}(\cdot)$, i.e., ${\bm{\sigma}} _{{\bm{\delta}}|{\bm{h}}} = f_{{\theta}_{gd}}(\bm{h})$. For the recovered label distribution $\bm{\hat{d}}$, we introduce the label distribution network $f_{{\theta}_{ld}}(\cdot)$ and has the following formula:
\begin{equation}
	\bm{\hat{d}} = f_{{\theta}_{ld}}(\bm{h}).
	\label{opt_end}
\end{equation}

Consequently, based on Eq.~(\ref{opt_start})-(\ref{opt_end}), we have:
\begin{equation}
	\resizebox{.887\hsize}{!}{$
	\begin{gathered}
		~~{\kern 1pt}~~ \mathop {\min }\limits_{{\theta _{en}},{\theta _{de}},{\theta _{gd}},{\theta _{ld}}} {{\mathcal{L}}_{LIB}} \hfill \\
		\Rightarrow \mathop {\min }\limits_{{\theta _{en}},{\theta _{de}},{\theta _{gd}},{\theta _{ld}}} \frac{1}{n} \sum\limits_{\bm{l}} [\frac{1}{2} {\left\| {{{\bm{\mu}} _{{\bm{l}}|{\bm{h}}}} - {\bm{l}}} \right\|_2^2}  \hfill \\
		+ \alpha (\frac{1}{2}{({\bm{l}} - \hat{{\bm{d}}})^T}({{\bm{\sigma}} _{{\bm{\delta}}|{\bm{h}}}^{-2}\bm{I}})({\bm{l}} - \hat{{\bm{d}}}) + \log \det ({{\bm{\sigma}} _{{\bm{\delta}}|{\bm{h}}}^2\bm{I}}))] \hfill \\ 
		+ \frac{\beta }{2}\sum\limits_{\bm{x}}[{{\bm{\mu}} _{{\bm{h}}|{\bm{x}}}^T}{{{\bm{\mu}} _{{\bm{h}}|{\bm{x}}}}} + {\rm{tr}}({{\bm{\sigma}} _{{\bm{h}}|{\bm{x}}}^2\bm{I}}) - \log \det ({{\bm{\sigma}} _{{\bm{h}}|{\bm{x}}}^2\bm{I}})]. \hfill \\
	\end{gathered} $}
\label{obj_perfect}
\end{equation}

When the problem of Eq.~(\ref{obj_perfect}) is optimized, we can effectively recover the desired label distributions. To be specific, given $\{\bm{X},\bm{L}\}$, we can obtain $\bm{H}$ according to Eq.~(\ref{obj_get_h}) and achieve the recovery results based on Eq.~(\ref{opt_end}), namely, $\bm{{\hat{D}}} = f_{{\theta}_{ld}}(\bm{H})$.

\subsection{Comparison with Existing LE Methods}

The main difference between LIB and existing methods is that our method deals with the problem of LE from the perspective of information bottleneck. Considering the first term in Eq.~(\ref{obj_basic_existing_methods}), it aims to minimize $\left\| {\bm{d} - \bm{l}} \right\|_2^2$ under the assumption that information in the
label distributions is inherited from the initial logical labels \cite{xuLETKDE,xu2020variational,tang2020label}. For LIB, the more reasonable term:
\begin{equation}
\frac{1}{2}{({\bm{l}} - \hat{{\bm{d}}})^T}({{\bm{\sigma}} _{{\bm{\delta}}|{\bm{h}}}^{-2}\bm{I}})({\bm{l}} - \hat{{\bm{d}}}) + \log \det ({{\bm{\sigma}} _{{\bm{\delta}}|{\bm{h}}}^2\bm{I}})
\label{reasonable_gap}
\end{equation}
which can be deduced by excavating the label relevant information about the label gaps between logical labels and label distributions.

Besides, we compare our method with the recently proposed LEVI \cite{xu2020variational} further. Although the objectives of LEVI and LIB are somewhat similar in form, they are essentially different as follows: 1) LIB makes attempts from the perspective of information bottleneck, while LEVI from the view of variational inference; 2) The formulas of LEVI and LIB are just partially similar in form, since the variational inference is employed as the optimization tool in LIB. The details of these two formulas are totally different; 3) LEVI requires an extra regularizer, i.e., $\left\| {\bm{d} - \bm{l}} \right\|_2^2$, to constrain the recovery process, while LIB achieves $\bm{d}$ based on the more reasonable term, i.e., Eq.~(\ref{reasonable_gap}).

\section{Experiments}
\label{sec:experiments}

To verify the effectiveness and competitiveness of LIB, extensive experiments are conducted in this section.

\begin{table}
	\centering
	\resizebox{.8 \columnwidth}!{
	\begin{tabular}{r|c|c|c}
		\toprule
		Dataset  & \# dimension $q$ & \# instance $n$ & \# labels $c$\\
		\midrule
		Artificial\_toy & 3 & 2601 & 3\\
		Movie & 1869 & 7755 & 5\\
		SBU-3DFE & 243 & 2500 & 6\\
		SJAFFE & 243 & 213 & 6\\
		Yeast-alpha & 24 & 2465 & 18\\
		Yeast-cdc & 24 & 2465 & 15\\
		Yeast-cold & 24 & 2465 & 4\\
		Yeast-diau & 24 & 2465 & 7\\
		Yeast-dtt & 24 & 2465 & 4\\
		Yeast-elu & 24 & 2465 & 14\\
		Yeast-heat & 24 & 2465 & 6\\
		Yeast-spo & 24 & 2465 & 6\\
		Yeast-spo5 & 24 & 2465 & 3\\
		Yeast-spoem & 24 & 2465 & 2\\
		\bottomrule
	\end{tabular}}
	\caption{Details of datasets. The numbers of dimension $q$, instance $n$, and labels $c$ are provided here.}
	\label{tab:datasets}
\end{table} 

\subsection{Experimental Setup}

As shown in Table \ref{tab:datasets}, we use both one toy dataset and 13 real-world datasets for evaluation\footnote{http://palm.seu.edu.cn/xgeng/LDL/index.htm}. For the toy dataset, i.e., Artificial dataset, it is utilized to vividly show the recovery performance \cite{xuLETKDE}. Movie dataset is collected from movies, SBU-3DFE and SJAFFE datasets are two facial expression datasets. Yeast datasets (alpha to spoem) are collected from 10 biological experiments on the budding yeast genes \cite{eisen1998cluster}. It is important to note that only the ground-truth label distributions are provided by these datasets. Therefore, we adopt the binarization strategy, which is also used in existing LE works \cite{xuLETKDE,xu2020variational,tang2020label}, to ensure the consistency of evaluation.

We compare our method LIB to 7 LE methods, including FCM \cite{FCM_LE}, KM \cite{KM_LE}, LP \cite{LP_LE}, ML \cite{ML_LE}, GLLE \cite{xuLETKDE}, LESC \cite{tang2020label}, and LEVI \cite{xu2020variational}. The first two methods belong to the algorithm adaption, and the rest methods are specialized algorithms. For the sake of fairness, we utilize the parameter settings recommended in their original works. Specifically, for FCM, we set the parameter $\beta = 2$. For KM, we leverage the Gaussian kernel. For LP, we set  the parameter $\alpha=0.5$. For ML, we set the number of neighbors $k=c+1$. For GLLE, we select $\lambda$ from $\{0.01,0.1,...,100\}$ and set the number of neighbors $k$ to $c+1$. For LESC, $\lambda_{1}$ and $\lambda_{2}$ are selected from $\{0.0001,0.1,...,10\}$. For LEVI, MLPs with two hidden layers and softplus activation functions are utilized, and the results are reported after 150 training epochs. For LIB, we select $\alpha$ and $\beta$ from $\{0.001,0.01,...,10\}$, and the fully connected networks with 3 layers and sigmoid activation function are leveraged in the proposed method. 

To evaluate the recovery performance, we adopt 6 metrics, namely Chebyshev, Canberra, Clark, Kullback-Leibler, Cosine, and Intersection \cite{TKDE2016_LDL_reivew,xuLETKDE,tang2020label}. Given the ground-truth label distribution $\bm{d}$ and the recovered label distribution $\bm{\hat{d}}$, the first four metrics and the rest two metrics respectively measure the distance and similarity between $\bm{d}$ and $\bm{\hat{d}}$:
\begin{gather}
	D_{\rm{Chebyshev}}(\bm{d},\bm{\hat{d}}) ={{\max }_{i}}\left| {{{d}}^{{{y}_{i}}}}-{{{{\hat{d}}}}^{{{y}_{i}}}} \right|,\hfill \\
	D_{\rm{Canberra}}(\bm{d},\bm{\hat{d}}) = \sum\limits_{i=1}^{c}{\frac{\left| {{d}^{{{y}_{i}}}}-{{{\hat{d}}}^{{{y}_{i}}}} \right|}{{{d}^{{{y}_{i}}}}+{{{\hat{d}}}^{{{y}_{i}}}}}}, \hfill \\
	D_{\rm{Clark}}(\bm{d},\bm{\hat{d}}) = \sqrt{\sum\limits_{i=1}^{c}{\frac{{{\left( {{d}^{{{y}_{i}}}}-{{{\hat{d}}}^{{{y}_{i}}}} \right)}^{2}}}{{{\left( {{d}^{{{y}_{i}}}}+{{{\hat{d}}}^{{{y}_{i}}}} \right)}^{2}}}}}, \hfill \\
	D_{\rm{Kullback-Leibler}}(\bm{d},\bm{\hat{d}}) = \sum\limits_{i = 1}^c {{d^{{y_i}}}\ln \frac{{{d^{{y_i}}}}}{{{{\hat d}^{{y_i}}}}}}, \hfill \\ 
	S_{\rm{Cosine}}(\bm{d},\bm{\hat{d}}) = \frac{\sum\limits_{i=1}^{c}{{{d}^{{{y}_{i}}}}}{{{\hat{d}}}^{{{y}_{i}}}}}{\sqrt{\sum\limits_{i=1}^{c}{{{\left( {{d}^{{{y}_{i}}}} \right)}^{2}}}}\sqrt{\sum\limits_{i=1}^{c}{{{\left( {{{\hat{d}}}^{{{y}_{i}}}} \right)}^{2}}}}}, \hfill \\ 
	S_{\rm{Intersection}}(\bm{d},\bm{\hat{d}}) = \sum\limits_{i=1}^{c}{\min }\left( {{d}^{{{y}_{i}}}},{{{\hat{d}}}^{{{y}_{i}}}}\right). 
\end{gather}
The smaller values of distance metric and similarity metric indicate the better and the worse results, respectively.

\subsection{Visualization Results on Toy Dataset}

\begin{figure}[t]
	\centering
	\begin{subfigure}{0.325\linewidth}
		\centering
		\includegraphics[width=1\linewidth]{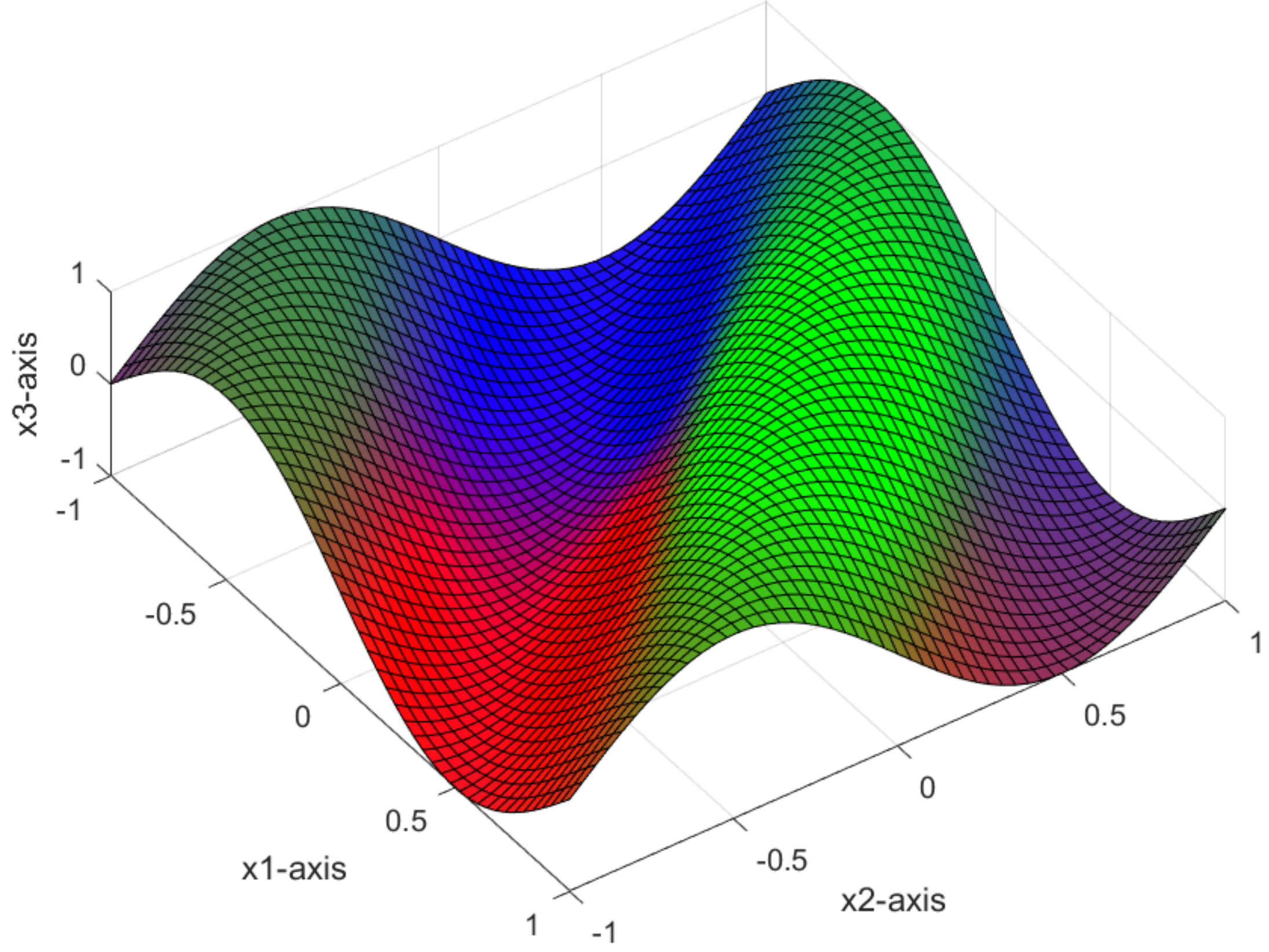}
		\caption{FCM}
		\label{FCM}
	\end{subfigure}
	\centering
	\begin{subfigure}{0.325\linewidth}
		\centering
		\includegraphics[width=1\linewidth]{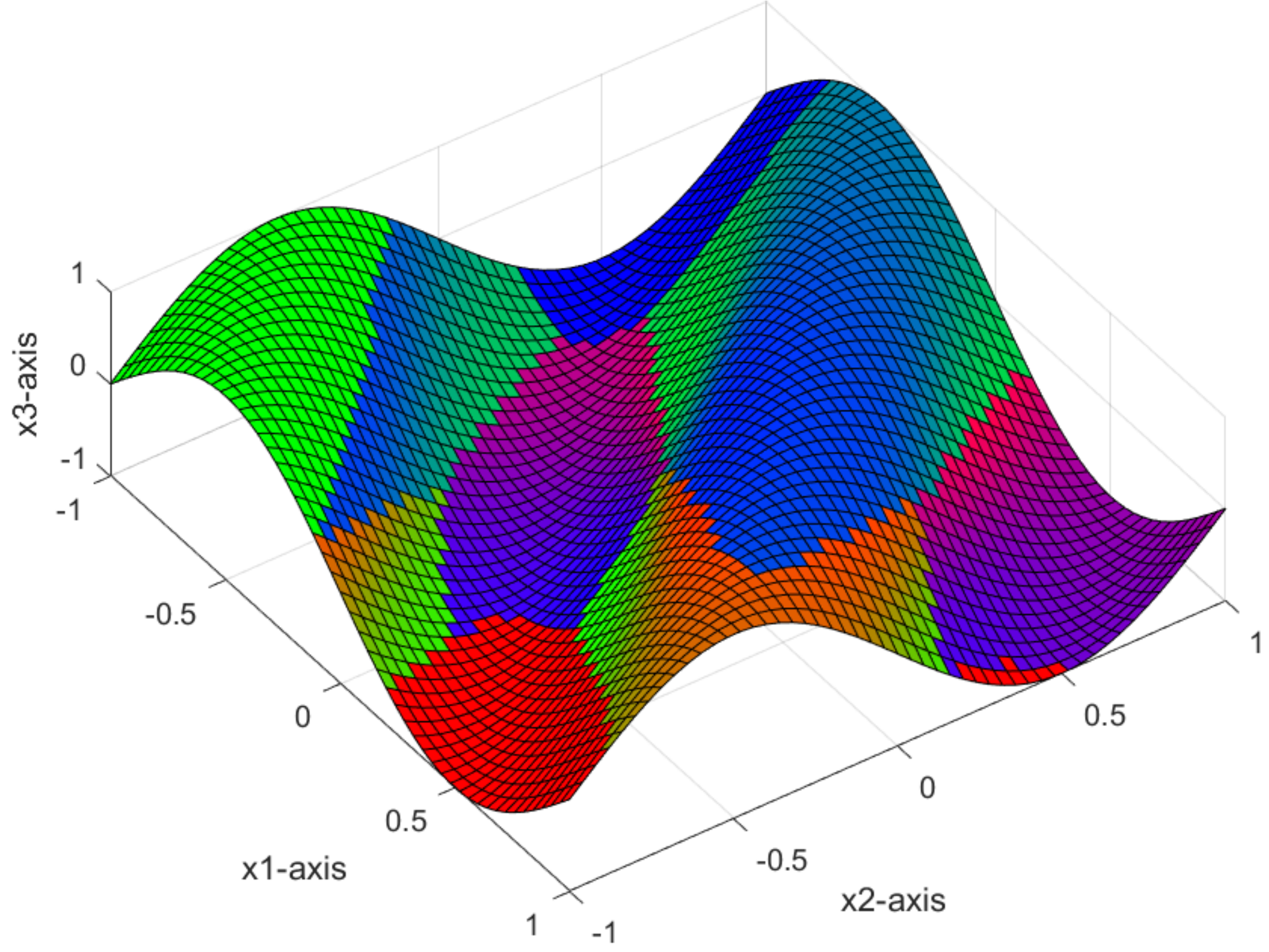}
		\caption{KM}
		\label{KM}
	\end{subfigure}
	\centering
	\begin{subfigure}{0.325\linewidth}
		\centering
		\includegraphics[width=1\linewidth]{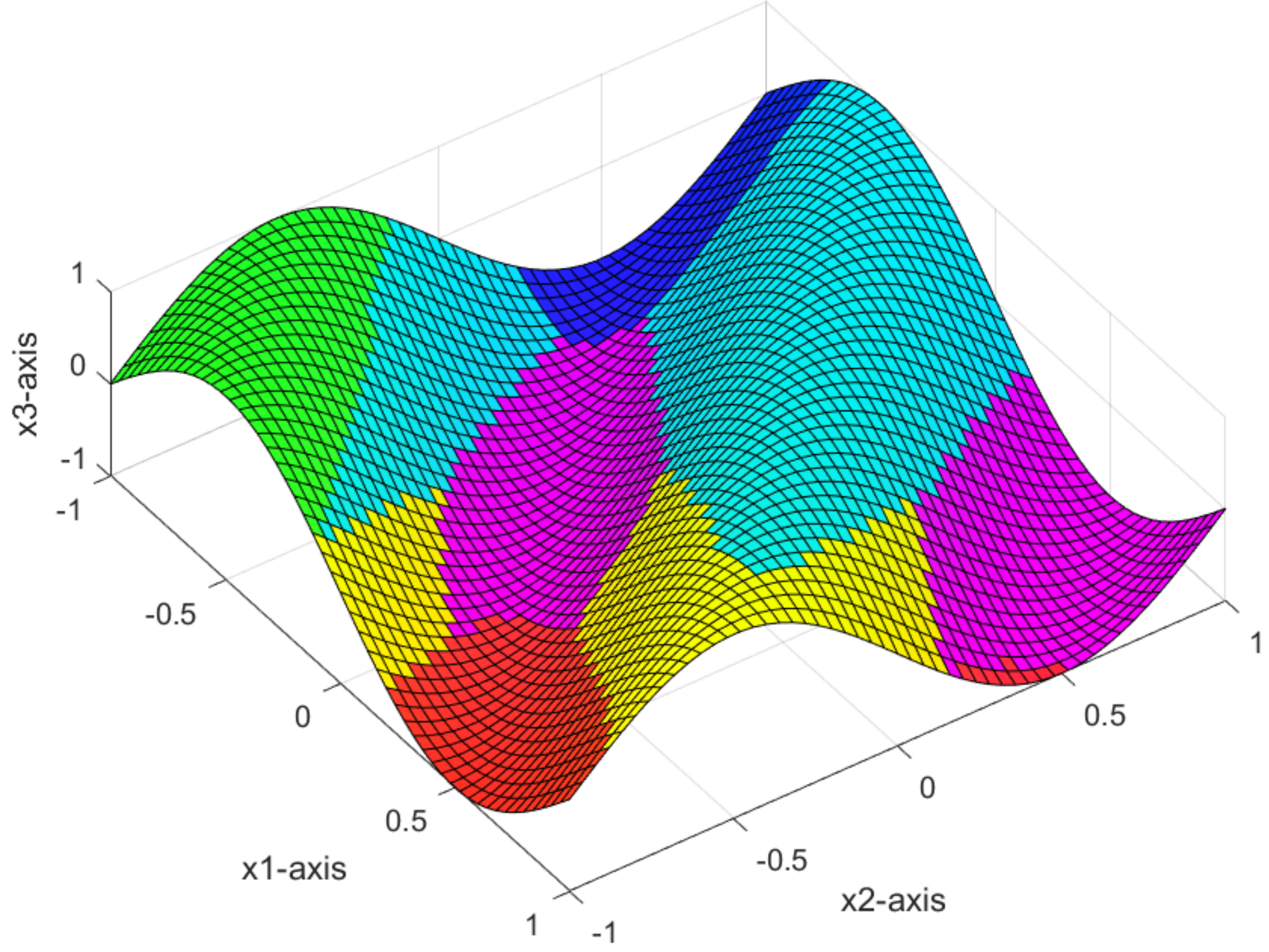}
		\caption{LP}
		\label{LP}
	\end{subfigure}
	\centering
	\begin{subfigure}{0.325\linewidth}
		\centering
		\includegraphics[width=1\linewidth]{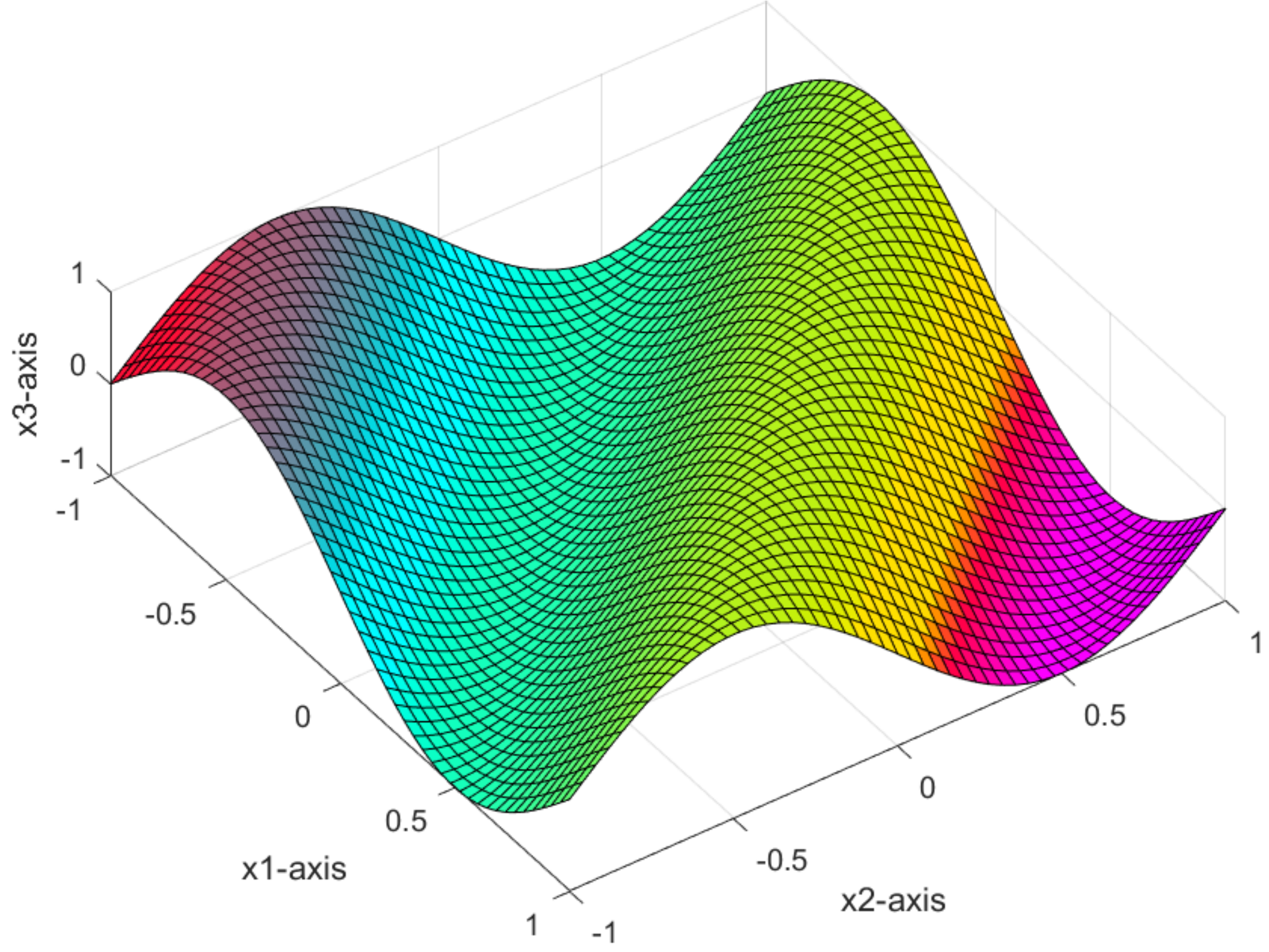}
		\caption{ML}
		\label{ML}
	\end{subfigure}
	\centering
	\begin{subfigure}{0.325\linewidth}
		\centering
		\includegraphics[width=1\linewidth]{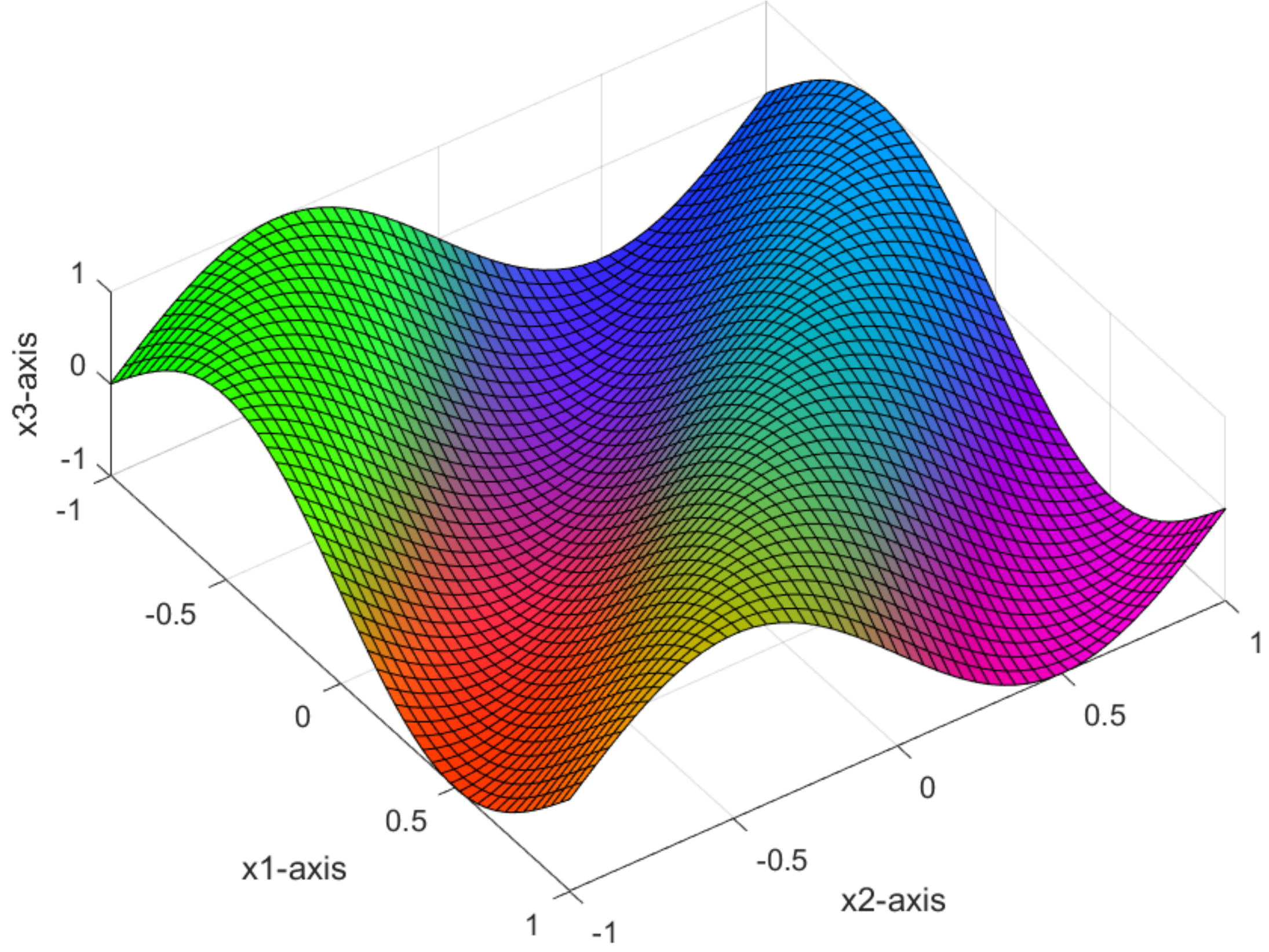}
		\caption{GLLE}
		\label{GLLE}
	\end{subfigure}
	\centering
	\begin{subfigure}{0.325\linewidth}
		\centering
		\includegraphics[width=1\linewidth]{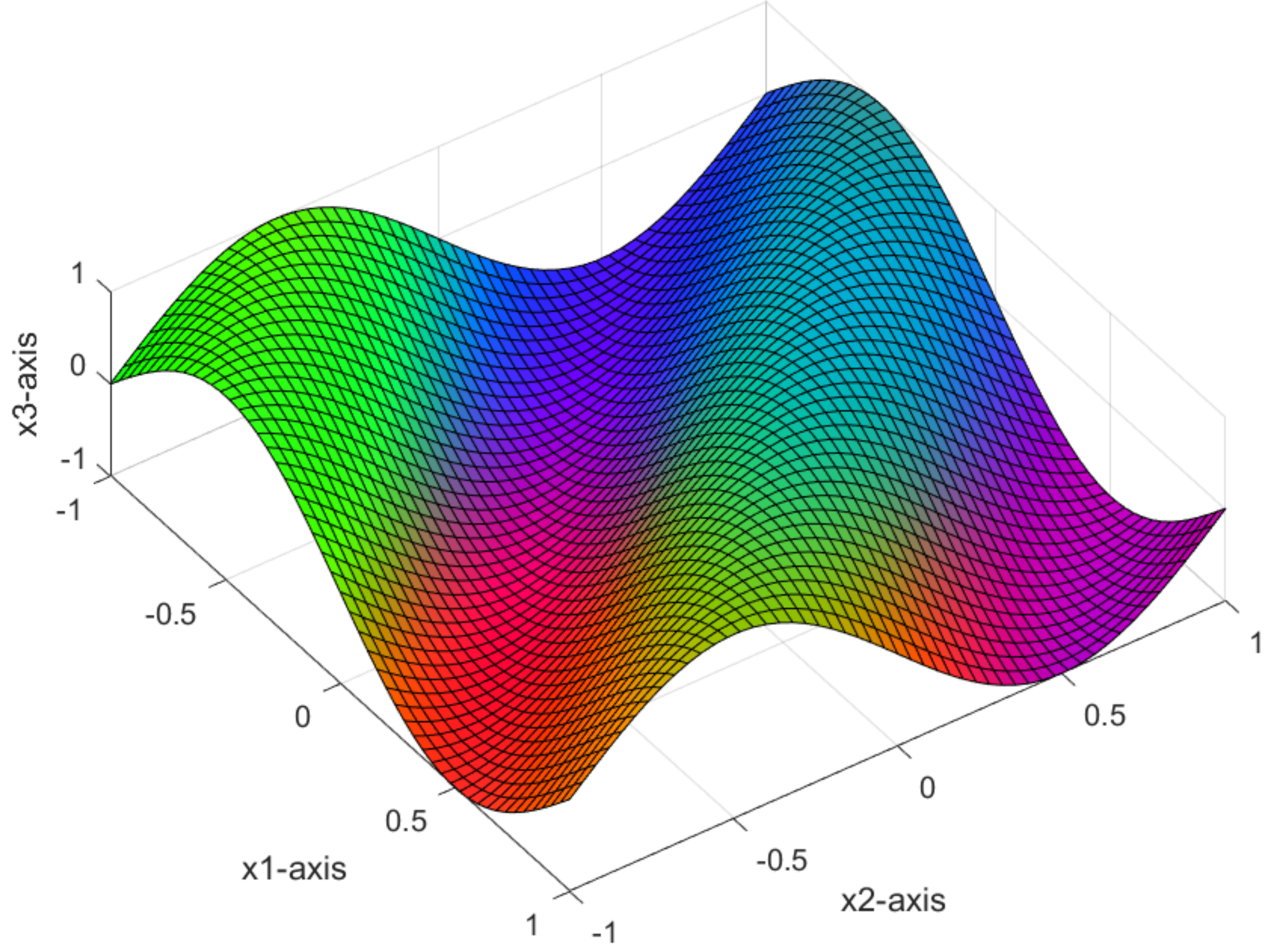}
		\caption{LESC}
		\label{LESC}
	\end{subfigure}
	\centering
	\begin{subfigure}{0.325\linewidth}
		\centering
		\includegraphics[width=1\linewidth]{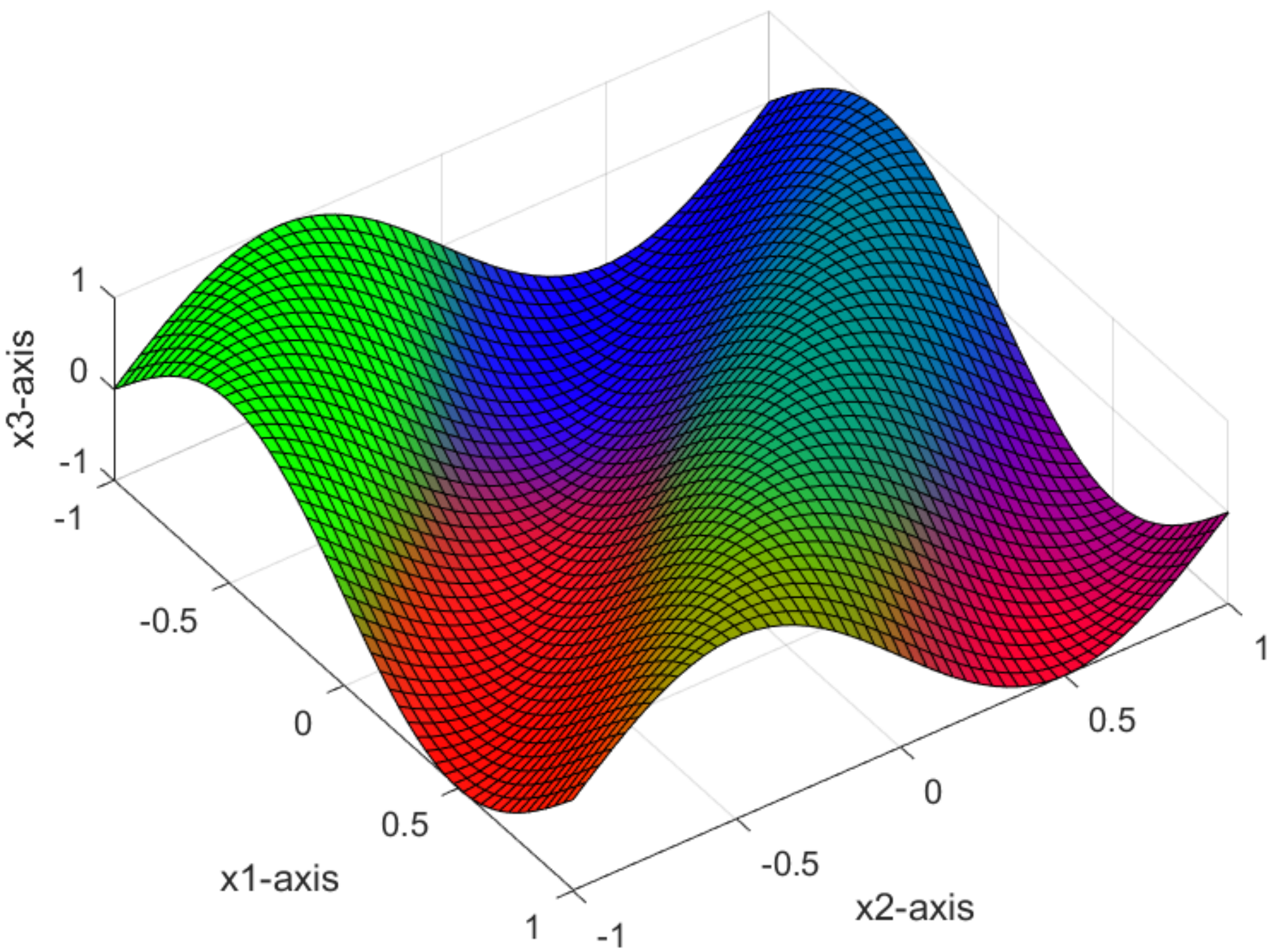}
		\caption{LEVI}
		\label{LEVI }
	\end{subfigure}
	\centering
	\begin{subfigure}{0.325\linewidth}
		\centering
		\includegraphics[width=1\linewidth]{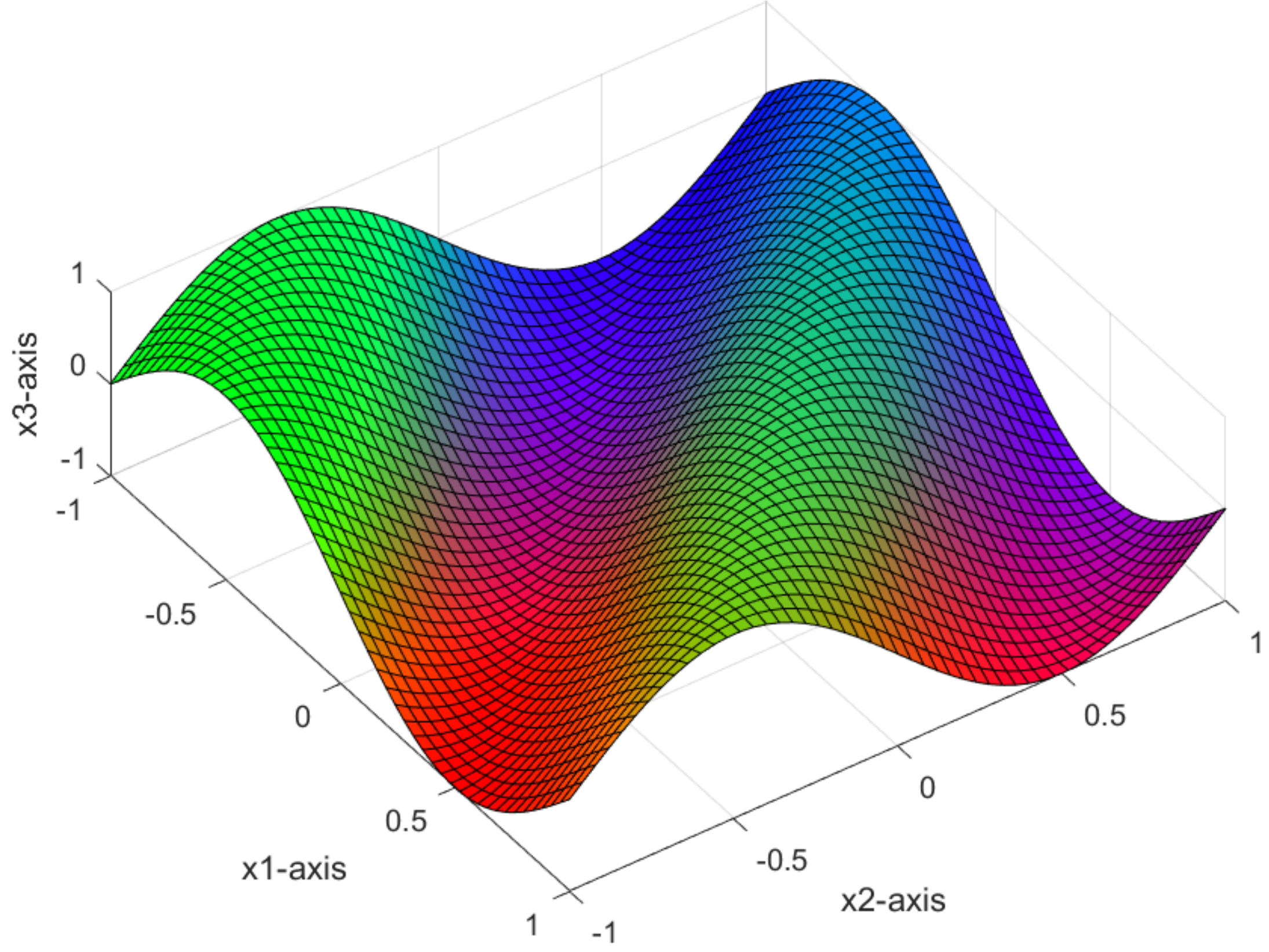}
		\caption{LIB}
		\label{LIB}
	\end{subfigure}
	\centering
	\begin{subfigure}{0.325\linewidth}
		\centering
		\includegraphics[width=1\linewidth]{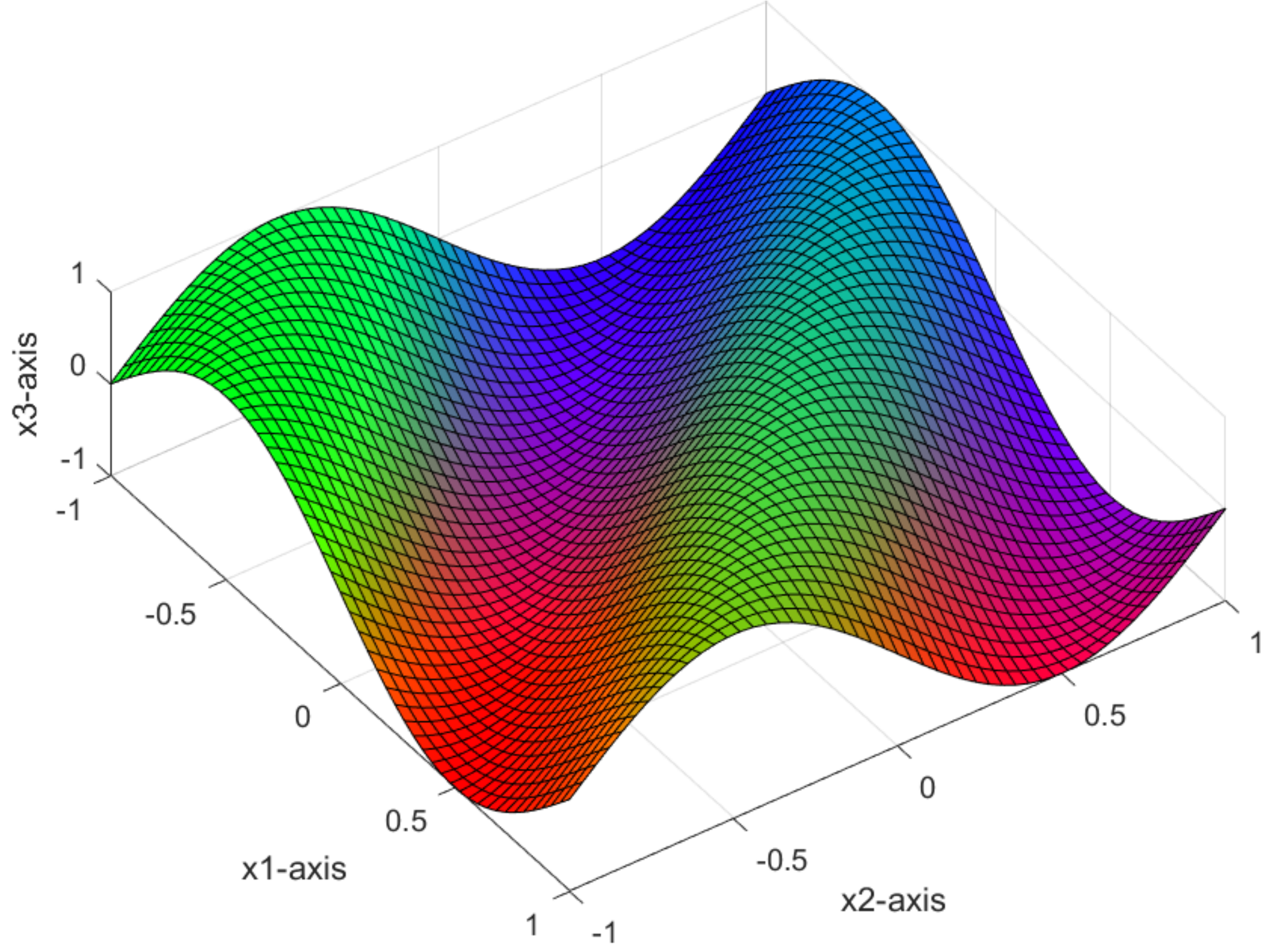}
		\caption{Ground-truth}
		\label{GT}
	\end{subfigure}
	\caption{Visualization results of the label distributions recovered by different methods ((a)-(h)) and the ground-truth ((i)) on Artificial dataset. (Best viewed in color.)}
	\label{fig:artificial_visualization}
\end{figure}

The recovery results on the Artificial dataset are vividly presented in Fig.~\ref{fig:artificial_visualization}, which shows the three-dimensional label distributions by the RGB color channels separately. The more similar the color patterns of the recovered results and the ground-truth are, the better the recovery results are.

It can be seen that FCM, GLLE, LESC, LEVI, and LIB can obtain the similar color pattern, while KM, LP, and ML are incapable to obtain the promising recovery performance on Artificial dataset. Regarding the visualization results of FCM, GLLE, LESC, LEVI, and LIB, the color pattern that is most close to the ground-truth is achieved by our LIB. 

\subsection{Comparison Results on Real-world Datasets}

\begin{table*}[htbp]
	\centering
	\resizebox{0.97\textwidth}!{
	\begin{tabular}{r|cccccccc|cccccccc}
		\toprule
		Metric & \multicolumn{8}{c|}{Chebyshev $\downarrow $}                         & \multicolumn{8}{c}{Clark $\downarrow $} \\
		\midrule
		Method & FCM   &  KM   &  LP   &  ML   &  GLLE &  LESC & LEVI & LIB   & FCM   &  KM   &  LP   &  ML   &  GLLE &  LESC & LEVI & LIB \\
		\midrule
		Movie & 0.230  & 0.234  & 0.161  & 0.164  & 0.122  & 0.121 & 0.110 & \textbf{0.107}  & 0.859  & 1.766  & 0.913  & 1.140  & 0.569  & 0.564 & 0.551  & \textbf{0.517}  \\
		SUB-3DFE & 0.135  & 0.238  & 0.123  & 0.233  & 0.126  & 0.122 & 0.095 & \textbf{0.094}  & 0.482  & 1.907  & 0.580  & 1.848  & 0.391  & 0.378 & 0.303  & \textbf{0.297}  \\
		SJAFFE & 0.132  & 0.214  & 0.107  & 0.186  & 0.087  & \textbf{0.069} & 0.075 & 0.071  & 0.522  & 1.874  & 0.502  & 1.519  & 0.377  & 0.276 & 0.290 & \textbf{0.262}  \\
		Yeast-alpha & 0.044  & 0.063  & 0.040  & 0.057  & 0.020  & 0.015 & \textbf{0.012} & 0.017  & 0.821  & 3.153  & 1.185  & 3.088  & 0.337  & \textbf{0.253} & 0.319 & 0.275  \\
		Yeast-cdc & 0.051  & 0.076  & 0.042  & 0.071  & 0.022  & 0.019 & \textbf{0.016} & 0.017 & 0.739  & 2.885  & 1.014  & 2.825  & 0.306  & 0.251 & 0.323 & \textbf{0.242}  \\
		Yeast-cold & 0.141  & 0.252  & 0.137  & 0.242  & 0.066  & 0.056 & 0.082 & \textbf{0.054}  & 0.433  & 1.472  & 0.503  & 1.440  & 0.176  & 0.152 & 0.269 & \textbf{0.146}  \\
		Yeast-diau & 0.124  & 0.152  & 0.099  & 0.148  & 0.053  & \textbf{0.042} & 0.044 & 0.049  & 0.838  & 1.886  & 0.788  & 1.844  & 0.296  & \textbf{0.224} & 0.295 & 0.273  \\
		Yeast-dtt & 0.097  & 0.257  & 0.128  & 0.244  & 0.052  & 0.043 & 0.084 & \textbf{0.034}  & 0.329  & 1.477  & 0.499  & 1.446  & 0.143  & 0.119 & 0.294 & \textbf{0.092}  \\
		Yeast-elu & 0.052  & 0.078  & 0.044  & 0.072  & 0.023  & 0.019 &  \textbf{0.017} & 0.018  & 0.579  & 2.768  & 0.973  & 2.711  & 0.295  & 0.241 & 0.317 & \textbf{0.224}  \\
		Yeast-heat & 0.169  & 0.175  & 0.086  & 0.165  & 0.049  & 0.046 & 0.052 & \textbf{0.039}  & 0.580  & 1.802  & 0.568  & 1.764  & 0.213  & 0.199 & 0.288 & \textbf{0.165}  \\
		Yeast-spo & 0.130  & 0.175  & 0.090  & 0.171  & 0.062  & 0.060 & 0.055 & \textbf{0.053}  & 0.520  & 1.811  & 0.558  & 1.768  & 0.266  & 0.258 & 0.277 & \textbf{0.224}  \\
		Yeast-spo5 & 0.162  & 0.277  & 0.114  & 0.273  & 0.099  & 0.092 & 0.091 & \textbf{0.076}  & 0.395  & 1.059  & 0.274  & 1.036  & 0.197  & 0.185 & 0.209 & \textbf{0.158}  \\
		Yeast-sopem & 0.233  & 0.408  & 0.163  & 0.403  & 0.088  & 0.087 & 0.115 & \textbf{0.069}  & 0.401  & 1.028  & 0.272  & 1.004  & 0.132  & 0.129 & 0.182 & \textbf{0.104}  \\
		\midrule
		Avg.Rank & 6.077  & 8.000  & 5.000  & 6.846  & 3.769  & 2.308 & 2.463  & \textbf{1.538}  & 5.385  & 8.000  & 5.615  & 7.000  & 3.385  & 1.923 & 3.462 & \textbf{1.231}  \\
		\midrule
		Metric & \multicolumn{8}{c}{Canberra $\downarrow $}                          & \multicolumn{8}{c}{Kullback-Leibler $\downarrow $} \\
		\midrule
		Method  & FCM   &  KM   &  LP   &  ML   &  GLLE &  LESC & LEVI & LIB   & FCM   &  KM   &  LP   &  ML   &  GLLE &  LESC & LEVI & LIB \\
		\midrule
		Movie & 1.664  & 3.444  & 1.720  & 1.934  & 1.045  & 1.034 & 0.974  & \textbf{0.920}  & 0.381  & 0.452  & 0.177  & 0.218  & 0.123  & 0.120 & 0.082  & \textbf{0.077}  \\
		SUB-3DFE & 1.020  & 4.121  & 1.245  & 4.001  & 0.820  & 0.799 & 0.637 & \textbf{0.611}  & 0.094  & 0.603  & 0.105  & 0.565  & 0.069  & 0.064 & 0.042 & \textbf{0.041}  \\
		SJAFFE & 1.081  & 4.010  & 1.064  & 3.138  & 0.781  & 0.561 & 0.600  & \textbf{0.531}  & 0.107  & 0.558  & 0.077  & 0.391  & 0.050  & 0.029 & 0.032 & \textbf{0.027}  \\
		Yeast-alpha & 2.883  & 11.809  & 4.544  & 11.603  & 1.134  & \textbf{0.846} & 1.249  & 0.893  & 0.100  & 0.630  & 0.121  & 0.602  & 0.013  & \textbf{0.008} & 0.011 & 0.009  \\
		Yeast-cdc & 2.415  & 9.875  & 3.644  & 9.695  & 0.959  & 0.765 & 1,148  & \textbf{0.747}  & 0.091  & 0.630  & 0.111  & 0.601  & 0.014  & 0.010 & 0.014 & \textbf{0.008}  \\
		Yeast-cold & 0.734  & 2.566  & 0.924  & 2.519  & 0.305  & 0.263 & 0.501  & \textbf{0.250}  & 0.113  & 0.586  & 0.103  & 0.556  & 0.019  & 0.015 & 0.035 & \textbf{0.012}  \\
		Yeast-diau & 1.895  & 4.261  & 1.748  & 4.180  & 0.671  & \textbf{0.480}  & 0.689 & 0.621  & 0.159  & 0.538  & 0.127  & 0.509  & 0.027  &\textbf{ 0.017} & 0.023 & 0.022  \\
		Yeast-dtt & 0.501  & 2.594  & 0.941  & 2.549  & 0.248  & 0.206 & 0.562  & \textbf{0.158}  & 0.065  & 0.617  & 0.103  & 0.586  & 0.013  & 0.010 & 0.042 & \textbf{0.005}  \\
		Yeast-elu & 1.689  & 9.110  & 3.381  & 8.949  & 0.902  & 0.727 & 1.093 & \textbf{0.670}  & 0.059  & 0.617  & 0.109  & 0.589  & 0.013  & 0.009 & 0.014 & \textbf{0.008}  \\
		Yeast-heat & 1.157  & 3.849  & 1.293  & 3.779  & 0.430  & 0.401 & 0.646 & \textbf{0.327}  & 0.147  & 0.586  & 0.089  & 0.556  & 0.017  & 0.015 & 0.027 & \textbf{0.011}  \\
		Yeast-spo & 0.998  & 3.854  & 1.231  & 3.772  & 0.548  & 0.533 & 0.605 & \textbf{0.454}  & 0.110  & 0.562  & 0.084  & 0.532  & 0.029  & 0.028 & 0.025 & \textbf{0.019}  \\
		Yeast-spo5 & 0.563  & 1.382  & 0.401  & 1.355  & 0.305  & 0.284 & 0.311 &\textbf{ 0.241}  & 0.123  & 0.334  & 0.042  & 0.317  & 0.034  & 0.031 & 0.028  & \textbf{0.021}  \\
		Yeast-sopem & 0.534  & 1.253  & 0.365  & 1.226  & 0.183  & 0.180 & 0.248  & \textbf{0.144}  & 0.208  & 0.531  & 0.067  & 0.503  & 0.027  & 0.027 & 0.036 & \textbf{0.018}  \\
		\midrule
		Avg.Rank & 5.231  & 8.000  & 5.692  & 7.000  & 3.231  & 2.00 & 3.692 & \textbf{1.154}  & 5.692  & 8.000  & 5.385  & 6.923  & 3.462  & 2.154 & 3.077  & \textbf{1.154}  \\
		\midrule
		Metric & \multicolumn{8}{c}{Cosine $\uparrow $}                            & \multicolumn{8}{c}{Intersection $\uparrow $} \\
		\midrule
		Method    & FCM   &  KM   &  LP   &  ML   &  GLLE &  LESC & LEVI & LIB   & FCM   &  KM   &  LP   &  ML   &  GLLE &  LESC & LEVI & LIB \\
		\midrule
		Movie & 0.773  & 0.880  & 0.929  & 0.919  & 0.936  & 0.937 & 0.954 & \textbf{0.955}  & 0.677  & 0.649  & 0.778  & 0.779  & 0.831  & 0.833 & 0.849  & \textbf{0.859}  \\
		SUB-3DFE & 0.912  & 0.812  & 0.922  & 0.815  & 0.927  & 0.932 & 0.956 & \textbf{0.958}  & 0.827  & 0.579  & 0.810  & 0.587  & 0.850  & 0.855 & 0.882 & \textbf{0.887}  \\
		SJAFFE & 0.906  & 0.827  & 0.941  & 0.857  & 0.958  & 0.973 & 0.969 & \textbf{0.974}  & 0.821  & 0.593  & 0.837  & 0.661  & 0.872  & 0.905 & 0.897 & \textbf{0.909}  \\
		Yeast-alpha & 0.922  & 0.751  & 0.911  & 0.756  & 0.987  & \textbf{0.992} & 0.989 & \textbf{0.992}  & 0.844  & 0.532  & 0.774  & 0.537  & 0.938  & \textbf{0.953} & 0.932 & 0.951  \\
		Yeast-cdc & 0.929  & 0.754  & 0.916  & 0.759  & 0.987  & 0.991 & 0.987  & \textbf{0.992}  & 0.847  & 0.533  & 0.779  & 0.538  & 0.937  & 0.950 & 0.925 & \textbf{0.951}  \\
		Yeast-cold & 0.922  & 0.779  & 0.925  & 0.784  & 0.982  & 0.986 & 0.970  &\textbf{ 0.988}  & 0.833  & 0.559  & 0.794  & 0.565  & 0.924  & 0.935 & 0.881 & \textbf{0.938}  \\
		Yeast-diau & 0.882  & 0.799  & 0.915  & 0.803  & 0.975  & \textbf{0.985} & 0.980 & 0.979  & 0.760  & 0.588  & 0.788  & 0.593  & 0.906  & \textbf{0.933} & 0.908 & 0.913  \\
		Yeast-dtt & 0.959  & 0.759  & 0.921  & 0.763  & 0.988  & 0.991 & 0.965  & \textbf{0.995}  & 0.894  & 0.541  & 0.786  & 0.546  & 0.939  & 0.949 & 0.866 & \textbf{0.961}  \\
		Yeast-elu & 0.950  & 0.758  & 0.918  & 0.763  & 0.987  & 0.991 & 0.987  & \textbf{0.992}  & 0.883  & 0.539  & 0.782  & 0.544  & 0.936  & 0.949 & 0.924 & \textbf{0.952}  \\
		Yeast-heat & 0.883  & 0.779  & 0.932  & 0.783  & 0.984  & 0.986 & 0.977 & \textbf{0.990}  & 0.807  & 0.559  & 0.805  & 0.564  & 0.929  & 0.934 & 0.897 & \textbf{0.946}  \\
		Yeast-spo & 0.909  & 0.800  & 0.939  & 0.803  & 0.974  & 0.975 & 0.978 & \textbf{0.982}  & 0.836  & 0.575  & 0.819  & 0.580  & 0.909  & 0.912 & 0.903 & \textbf{0.925}  \\
		Yeast-spo5 & 0.922  & 0.882  & 0.969  & 0.884  & 0.971  & 0.974 & 0.979  & \textbf{0.983}  & 0.838  & 0.724  & 0.886  & 0.727  & 0.901  & 0.908 & 0.909 & \textbf{0.924}  \\
		Yeast-sopem & 0.878  & 0.812  & 0.950  & 0.815  & 0.978  & 0.978 & 0.972 & \textbf{0.985}  & 0.767  & 0.592  & 0.837  & 0.597  & 0.912  & 0.913 & 0.885 & \textbf{0.931}  \\
		\midrule
		Avg.Rank & 5.846  & 7.923  & 5.308  & 6.923  & 3.462  & 2.154 & 2.923 & \textbf{1.231}  & 5.385  & 8.000  & 5.692  & 6.846  & 3.385  & 2.007 & 3.462 & \textbf{1.154}  \\
		\bottomrule
	\end{tabular}}
	\caption{Recovery results on 13 real-world datasets. $\downarrow$ indicates that ``the smaller the better'' and $\uparrow$ means that ``the larger the better''. The average ranks (Avg.Rank) on all datasets are also reported for all methods. We highlight the best recovery results.}
	\label{tab:comparison_results}
\end{table*}%

We provide the detailed comparison results on 13 real-world datasets in Table \ref{tab:comparison_results}. Overall, LIB has the competitive recovery performance. We have the following observations: 1) Compared with FCM and KM, which belong to the category of algorithm adaption, remarkable improvements can be achieved by our method; 2) Compared with the methods belonging to the category of specialized algorithm, LIB can also obtain better recovery results in most cases. For example, LIB obtains the best recovery results on Movie datasets in all metrics. Moreover, although LESC can obtain slightly favorable results in some cases, the corresponding recovery results of LIB are also promising and competitive. The underlying reason may be that LESC further considers the sample correlations during the recovery process; 3) The recovery performance of all methods can be roughly ranked as LIB$>$LESC$\approx$LEVI$>$GLLE$>$LP$\approx $FCM$>$ML$>$KM. We can conclude that LIB is suitable for the LE problem. The underlying reason for the significant improvement is that the label relevant information, including the information about label assignments and the information about label gaps, can be effectively investigated by LIB.

\subsection{Analysis and Discussion of LIB}

We analyze the parameter sensitivity of LIB firstly, and then we conduct the ablation studies as well. 

\subsubsection{Sensitivity of LIB}

\begin{figure*}[t]
	\centering
	\begin{subfigure}{0.29\linewidth}
		\centering
		\includegraphics[width=1\linewidth]{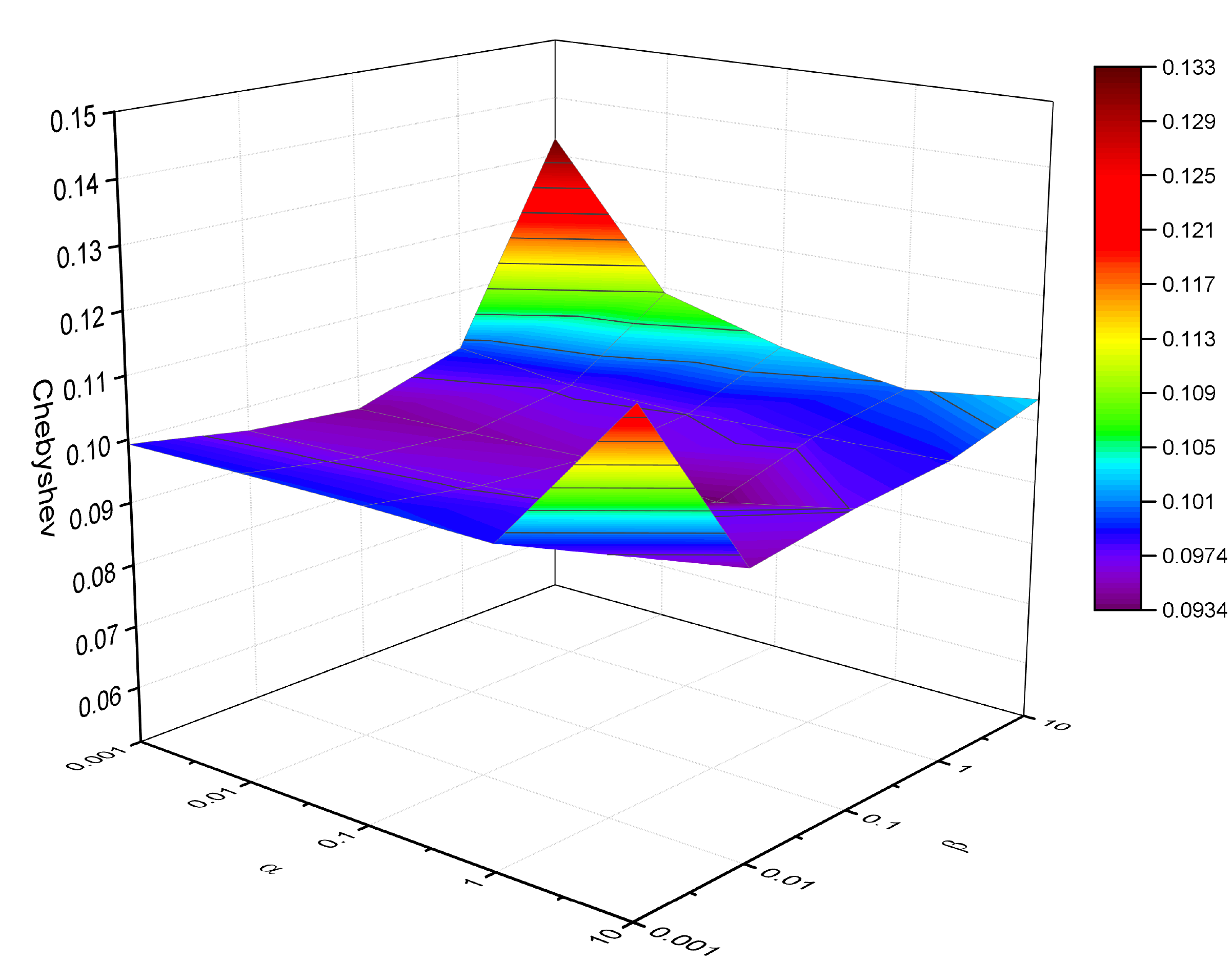}
		\caption{Chebyshev $\downarrow$}
		\label{Chebyshev}
	\end{subfigure}
	\centering
	\begin{subfigure}{0.29\linewidth}
		\centering
		\includegraphics[width=1\linewidth]{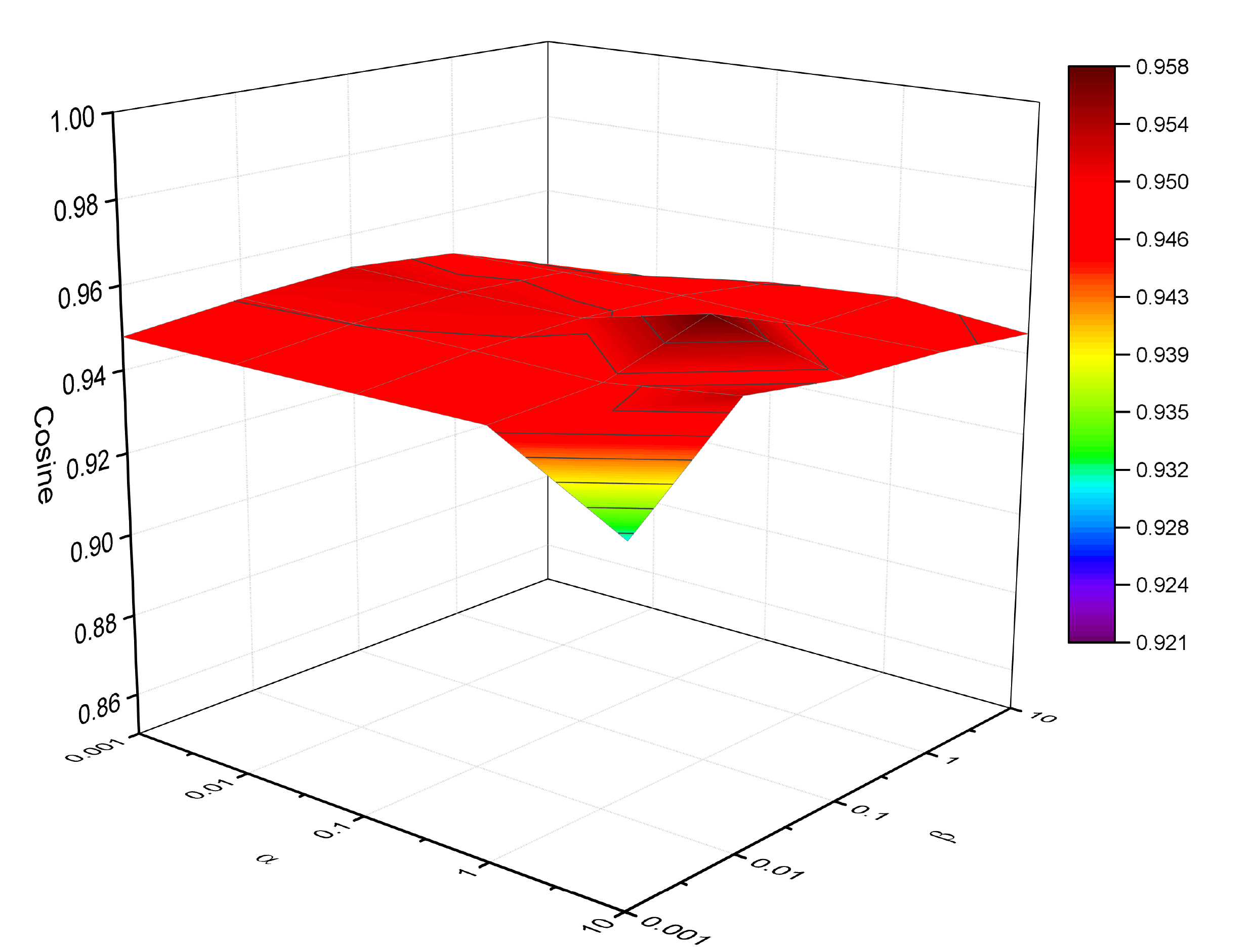}
		\caption{Cosine $\uparrow$}
		\label{Cosine}
	\end{subfigure}
	\centering
	\begin{subfigure}{0.29\linewidth}
		\centering
		\includegraphics[width=1\linewidth]{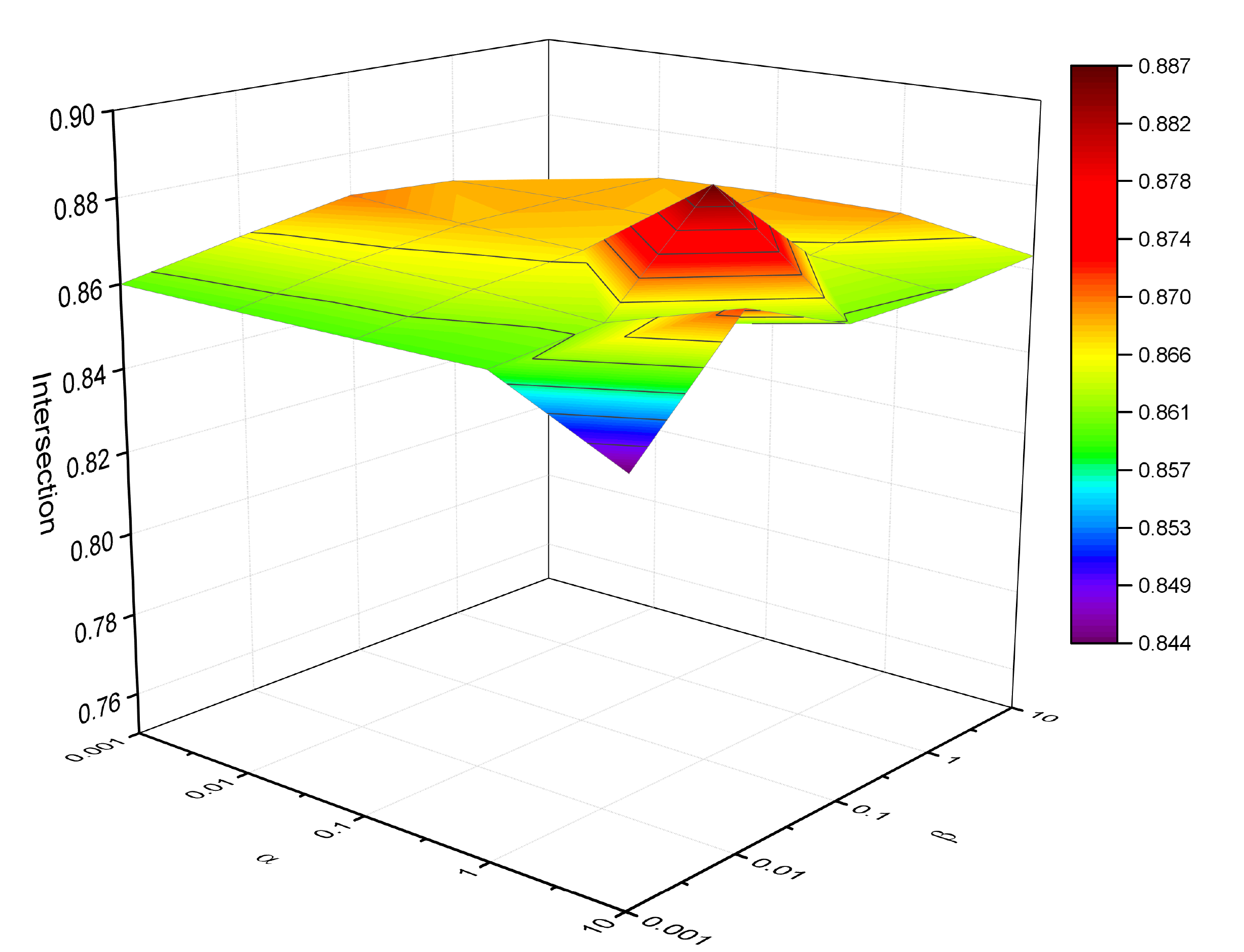}
		\caption{Intersection $\uparrow$}
		\label{Intersection}
	\end{subfigure}
	\caption{The recovery results in metrics of (a) Chebyshev distance, (b) Cosine coefficient, and (c) Intersection similarity with different values of $\alpha$ and $\beta$.  $\downarrow$ indicates that ``the smaller the better'' and $\uparrow$ means that ``the larger the better''. The experimental results demonstrate that LIB is robust with respect to different values of $\lambda$ and $\beta$. (Best viewed in color.)}
	\label{fig:para_tuning}
\end{figure*}

In the proposed method, we choose the values of $\alpha$ and $\beta$ from $\{0.001,0.01,...,10\}$. To show the parameter sensitivity of LIB, we conduct experiments on SBU-3DFE datasets with different values of $\alpha$ and $\beta$. Regarding the dimension of the learned latent representation, we set it to 256 for all datasets. The experimental results in metrics of Chebyshev distance, Cosine coefficient, and Intersection similarity are provided in Fig.~\ref{fig:para_tuning}. It can be observed that LIB method can get promising recovery results and is robust with respect to different values of $\alpha$ and $\beta$ in a large range. 

\begin{table*}[t]
	\centering
	\resizebox{0.87\textwidth}!{
	\begin{tabular}{r|cc|cc|cc|cc|cc|cc}
		\toprule
		Metric & \multicolumn{2}{c|}{Chebyshev $\downarrow$} & \multicolumn{2}{c|}{Clark $\downarrow$} & \multicolumn{2}{c|}{Canberra $\downarrow$} & \multicolumn{2}{c|}{Kullback-Leibler $\downarrow$} & \multicolumn{2}{c|}{Cosine $\uparrow$} & \multicolumn{2}{c}{Intersection $\uparrow$} \\
		\midrule
		Method & \multicolumn{1}{c}{LIB$_{gap}$} & \multicolumn{1}{c|}{LIB} & \multicolumn{1}{c}{LIB$_{gap}$} & \multicolumn{1}{c|}{LIB} & \multicolumn{1}{c}{LIB$_{gap}$} & \multicolumn{1}{c|}{LIB} & \multicolumn{1}{l}{LIB$_{gap}$} & \multicolumn{1}{c|}{LIB} & \multicolumn{1}{c}{LIB$_{gap}$} & \multicolumn{1}{c|}{LIB} & \multicolumn{1}{c}{LIB$_{gap}$} & \multicolumn{1}{c}{LIB} \\
		\midrule
		Movie & 0.120  & \textbf{0.107}  & 0.563  & \textbf{0.517}  & 1.029  & \textbf{0.920}  & 0.099  & \textbf{0.077}  & 0.938  & \textbf{0.955}  & 0.834  & \textbf{0.859}  \\
		SUB-3DFE & 0.130  & \textbf{0.094}  & 0.395  & \textbf{0.297}  & 0.849  & \textbf{0.611}  & 0.079  & \textbf{0.041}  & 0.923  & \textbf{0.958}  & 0.846  & \textbf{0.887}  \\
		SJAFFE & 0.113  & \textbf{0.071}  & 0.391  & \textbf{0.262}  & 0.816  & \textbf{0.531}  & 0.066  & \textbf{0.027}  & 0.938  & \textbf{0.973}  & 0.860  & \textbf{0.909}  \\
		Yeast-alpha & 0.018  & \textbf{0.017}  & 0.281  &\textbf{ 0.275}  & 0.920  & \textbf{0.893}  & {0.010}  & \textbf{0.009}  & 0.991  & \textbf{0.992}  & 0.950  &\textbf{ 0.951}  \\
		Yeast-cdc & 0.019  & \textbf{0.017}  & 0.254  & \textbf{0.242}  & 0.782  & \textbf{0.747}  & 0.009  & \textbf{0.008}  & 0.991  & \textbf{0.992}  & 0.948  & \textbf{0.951}  \\
		Yeast-cold & 0.061  & \textbf{0.017}  & 0.162  &\textbf{ 0.146}  & 0.280  & \textbf{0.250}  & 0.016  & \textbf{0.012}  & 0.985  & \textbf{0.988}  & 0.930  & \textbf{0.938}  \\
		Yeast-diau & 0.050  & \textbf{0.049}  & 0.288  & \textbf{0.273}  & 0.659  & \textbf{0.621}  & 0.025  & \textbf{0.022}  & 0.977  & \textbf{0.979}  & 0.908  & \textbf{0.913}  \\
		Yeast-dtt & 0.045  & \textbf{0.034}  & 0.124  & \textbf{0.092}  & 0.217  & \textbf{0.158}  & 0.010  & \textbf{0.005}  & 0.991  & \textbf{0.995}  & 0.946  & \textbf{0.961}  \\
		Yeast-elu & 0.019  & \textbf{0.018}  & 0.237  & \textbf{0.224}  & 0.714  & \textbf{0.670}  & 0.009  & \textbf{0.008}  & 0.992  & \textbf{0.992}  & 0.949  & \textbf{0.952}  \\
		Yeast-heat & 0.045  & \textbf{0.039}  & 0.193  & \textbf{0.165}  & 0.388  & \textbf{0.327}  & 0.014  & \textbf{0.011}  & 0.986  & \textbf{0.990}  & 0.936  & \textbf{0.946}  \\
		Yeast-spo & 0.059  & \textbf{0.053}  & 0.253  & \textbf{0.224}  & 0.523  & \textbf{0.454}  & 0.025  & \textbf{0.019}  & 0.976  & \textbf{0.982}  & 0.914  & \textbf{0.925}  \\
		Yeast-spo5 & 0.097  & \textbf{0.076}  & 0.193  & \textbf{0.158}  & 0.300  & \textbf{0.241}  & 0.032  & \textbf{0.021}  & 0.971  & \textbf{0.983}  & 0.903  & \textbf{0.924}  \\
		Yeast-sopem & 0.088  & \textbf{0.069}  & 0.130  & \textbf{0.104}  & 0.181  & \textbf{0.144}  & 0.027  & \textbf{0.018}  & 0.977  & \textbf{0.985}  & 0.912  & \textbf{0.931}  \\
		\bottomrule
	\end{tabular}}
	\caption{Recovery results of LIB$_{gap}$ and LIB on 13 real-world datasets. $\downarrow$ indicates that ``the smaller the better'' and $\uparrow$ means that ``the larger the better''. We highlight the best recovery results.}
	\label{tab:ablation}
\end{table*}

\subsubsection{Ablation studies of LIB}

The ablation studies are conducted to further verify the effectiveness of introducing the label information bottleneck framework for LE. In the proposed objective Eq.~(\ref{obj:lib_brief}), ${{\mathcal{L}}_{as}}$ and ${{\mathcal{L}}_{gap}}$ explore the label assignments information and label gaps information during the recovery process. As can be observed from Eq.~(\ref{obj:label_assignment_bound}) and Eq.~(\ref{obj:label_gaps_bound}), only the latent representation $\bm{H}$ can be learned if we employ ${{\mathcal{L}}_{as}}$ merely during the recovery process. Consequently, considering the goal of LE, we compare the proposed LIB with the method termed LIB$_{gap}$, which only employs ${{\mathcal{L}}_{gap}}$ to achieve the recovery results. In other words, LIB$_{gap}$ investigates the label gaps information in the case of not considering the label assignments information during the recovery process.

To be specific, LIB$_{gap}$ has with the following objective:
\begin{equation}
		\mathop {\min }\limits_{{\theta _{gd}},{\theta _{ld}}} \frac{1}{2} \sum\limits_{\bm{x}} [{({\bm{l}} - \hat{{\bm{d}}})^T}({{\bm{\sigma}} _{{\bm{\delta}}|{\bm{x}}}^{-2}\bm{I}})({\bm{l}} - \hat{{\bm{d}}})  + \log \det ({{\bm{\sigma}} _{{\bm{\delta}}|{\bm{x}}}^2\bm{I}})]. 
\end{equation}
Notably, ${\bm{\sigma}} _{{\bm{\delta}}|{\bm{x}}} = f_{{\theta}_{gd}}(\bm{x})$ and $\bm{\hat{d}} = f_{{\theta}_{ld}}(\bm{x})$, which are different from the objective Eq.~(\ref{obj_perfect}) utilized in LIB. Only the partial label relevant information, i.e., the information about the label gaps, is explored in
LIB$_{gap}$. 

Table \ref{tab:ablation} provides the recovery results of LIB$_{gap}$ and LIB. It can be observed that LIB outperforms LIB$_{gap}$ in all cases. Compared with LIB, LIB$_{gap}$ merely makes the effort to explore the label gap information to boost the recovery performance, while LIB excavates both the information about the label assignments and the information about the label gaps jointly. Therefore, the promising recovery performance can be achieved by LIB. Furthermore, according to the results provided in Table \ref{tab:comparison_results} and \ref{tab:ablation}, the recovery results of LIB$_{gap}$ seem to be competitive, which also indicates that the exploration of information about label gaps is beneficial for LE.

\section{Conclusion}
In this paper, we present a new perspective to deal with the Label Enhancement (LE) problem and introduce the novel Label Information Bottleneck (LIB) method. The label relevant information is decomposed into the information about label assignments and the information about label gaps. Consequently, our method transform the LE problem into simultaneously learning the latent representation and modeling the label gaps. Extensive experiments carried on both the toy dataset and real-world datasets verify the competitiveness of LIB.

\section*{Acknowledgments}
This work was supported by the National Key R\&D Program of China under Grant 2020AAA0109602; the Education and Research Foundation for Middle-aged and Young Teacher of Fujian Province under Grant JAT220005; the National Natural Science Foundation of China (NSFC) under Grant 62206156; Alibaba Group through Alibaba Innovative Research Program, No.21169774.   

{\small
\bibliographystyle{ieeefullname}
\bibliography{egbib}
}

\end{document}